\documentclass[11pt]{article}
\usepackage[centering,top=1.5in,bottom=1.2in,left=1.1in,right=1.1in]{geometry}
\usepackage[utf8]{inputenc}
\usepackage{verbatim}
\usepackage[sort,numbers]{natbib}
\usepackage{graphicx,subcaption}
\usepackage{graphbox}
\usepackage{amsmath,amssymb,amsthm}
\usepackage{hyperref}
\usepackage{bm}
\usepackage{enumitem}
\usepackage[usenames]{color}
\usepackage[dvipsnames]{xcolor}
\usepackage{listings}
\usepackage{algorithm,algorithmic}
\usepackage{eucal}
\usepackage{systeme}
\usepackage{caption} 
\usepackage{extarrows}
\usepackage{newtxmath}
\usepackage{palatino}
\usepackage{amsmath}
\usepackage{tikz}
\usetikzlibrary{arrows,shapes.geometric,automata}
\newtheorem{theorem}{Theorem}[section]
\newtheorem{lemma}[theorem]{Lemma}
\newtheorem{definition}[theorem]{Definition}
\newtheorem{example}[theorem]{Example}

\theoremstyle{remark}
\newtheorem*{remark}{Remark}
\newtheorem*{examplenn}{Example}

\def\R{\mathbb{R}}
\def\C{\mathbb{C}}
\def\F{\mathbb{F}}
\def\M{\mathcal{M}}
\def\dag{H}
\def\grad{\nabla}

\def\poly{\text{poly}}
\title{
Fast Global Convergence for Low-rank Matrix Recovery via Riemannian Gradient Descent with Random Initialization
}
\date{}
\author{\small Thomas Y. Hou, Zhenzhen Li, Ziyun Zhang}
\begin{document}
\maketitle
\graphicspath{{./figures/}}
\begin{abstract}
In this paper, we propose a new global analysis framework for a class of low-rank matrix recovery problems on the Riemannian manifold. We analyze the global behavior for the Riemannian optimization with random initialization. We use the Riemannian gradient descent algorithm to minimize a least squares loss function, and study the asymptotic behavior as well as the exact convergence rate. We reveal a previously unknown geometric property of the low-rank matrix manifold, which is the existence of spurious critical points for the simple least squares function on the manifold. We show that under some assumptions, the Riemannian gradient descent starting from a random initialization with high probability avoids these spurious critical points and only converges to the ground truth in nearly linear convergence rate, i.e. $\mathcal{O}(\text{log}(\frac{1}{\epsilon})+ \text{log}(n))$ iterations to reach an $\epsilon$-accurate solution. We use two applications as examples for our global analysis. The first one is a rank-1 matrix recovery problem. The second one is a generalization of the Gaussian phase retrieval problem. It only satisfies the weak isometry property, but has behavior similar to that of the first one except for an extra saddle set. Our convergence guarantee is nearly optimal and almost dimension-free, which fully explains the numerical observations. The global analysis can be potentially extended to other data problems with random measurement structures and empirical least squares loss functions.
\end{abstract}
\section{Introduction}\label{sec: intro}
Low-rank matrix recovery problems have been extensively studied in machine learning, signal processing, imaging science, advanced statistics, information theory and quantum mechanics, etc. Related problems include but are not limited to matrix factorization, matrix sensing, matrix completion, phase retrieval, robust PCA, density matrix detection, subspace clustering, to name a few. 

Earlier approaches on these problems include convex relaxation methods \cite{candes2009exact,candes2010power,recht2010guaranteed,candes2011tight,candes2013phaselift}, which convexify the problems by replacing rank minimization with nuclear norm minimization, and achieve exact recovery with provable guarantee. More recent works have shifted focus to nonconvex methods due to their lighter computational cost. Specifically, those methods mainly depend on the Burer-Monteiro factorization (e.g. parameterize a rank-r matrix $Z=(U,V)=UV^T$, with $U,V\in\F^{n\times r}$) and the global landscape analysis of the corresponding nonconvex objective function \cite{MaSen,MaCompletion,PhaseRetrieval,li2019toward,TenDecomp,DicLearn,robustPCA,ge2017no}. For example, for some specific machine learning problems, it has been shown that the landscape does not have spurious local minima. This property, combined with the convergence results in nonconvex optimization \cite{TenDecomp,Jinchi2017}, leads to convergence guarantee for second-order stationary points. We refer the reader to the excellent survey papers \cite{chi2019nonconvex,halko2011finding, recht2010guaranteed,hu2020brief} on low-rank matrix recovery.

Despite such progress, many questions remain unanswered. In particular, the convergence rate results obtained by those nonconvex methods are largely sub-linear, in sharp contrast to numerical observations indicating fast and nearly linear convergence. Furthermore, the Burer-Monteiro factorization \cite{burer2003nonlinear,burer2005local} may result in uncountably many artificial critical points. Lastly, individual problems are usually handled case by case without a full understanding of the intrinsic mechanism behind these problems.

To understand these questions, we propose a unified global analysis framework from the viewpoint of a Riemannian manifold. Instead of factorizing the low-rank matrix, we impose the low-rank constraint by restricting the domain to the low-rank matrix manifold $\mathcal{M}_r:=\{Z: \text{rank}(Z)=r, \, Z\in\F^{n_1\times n_2}\}$, $\F=\R$ or $\C$. We consider the following optimization problem over this low-rank manifold: 
\begin{equation}\label{eq: maniopt}
    \min_{Z\in\mathcal{M}_r}f(Z)=\frac{1}{2}\|T(Z)-y\|_2^2,
\end{equation}  
where $T:\M_r\rightarrow \R^m$ is a linear operator, $T(Z) = \frac{1}{\sqrt{m}}(\langle A_1,Z\rangle,\ldots, \langle A_m,Z\rangle)^\top$, and $y\in\R^m$ with $y_j = \frac{1}{\sqrt{m}}\langle A_j,X\rangle$. The formulation is general and covers many different low-rank matrix recovery problems, as shown in the following examples.
\begin{examplenn}
     We give a few specific examples of the operator $T$ and measurements $\{A_j\}_{j=1}^m$.
    \begin{enumerate}[label=\arabic*)]
    \item Matrix sensing: $T: \M_r \rightarrow\R^m$, where $\{A_j\}_{j=1}^m \subset \F^{n_1\times n_2}$ have entries drawn i.i.d from $\mathcal{N}(0,1)$, if $\F=\R$; and $\frac{\sqrt{2}}{2}\mathcal{N}(0,1)+i\frac{\sqrt{2}}{2}\mathcal{N}(0,1)$, if $\F=\C$. 
    \item Matrix completion: $T: \M_r \rightarrow\R^m$, where $\{A_j\}_{j=1}^m \subset \F^{n_1\times n_2}$ are generated by a uniform sampling of indices $\Omega\subset [n_1]\times[n_2]$ of a $n_1\times n_2$ matrix. The matrix $A_j$ is the indicator of the $j$-th sampled entry, i.e. with value 1 in the sampled indices and 0's in the other indices.  
    \item Gaussian phase retrieval: $
    T: \M_1 \rightarrow\R^m$, where $ \M_1$ is the symmetric rank-1 matrix manifold, and $\{A_j\}_{j=1}^m \subset \F^{n\times n}$ are rank-1 matrices. In the real case,  $A_j=a_ja_j^\top$ with $a_j\in\R^n$ and their entries are drawn i.i.d from $\mathcal{N}(0,1)$; in the complex case, $A_j=a_ja_j^*$ with $a_j\in\C^n$ and their entries are drawn i.i.d from $\frac{\sqrt{2}}{2}\mathcal{N}(0,1)+i\frac{\sqrt{2}}{2}\mathcal{N}(0,1)$.
\end{enumerate}
\end{examplenn}

All the above examples can be considered as random sensing of a low-rank matrix, where $T$ is a linear operator and $A_j$'s are drawn from some random distribution. There are also works on the number of measurements and their distributions in order to guarantee successful recovery. Despite the difference in the problem setting in these models and the distribution of $A_j$'s, their population problems share some common properties. For this reason, we will focus on the global convergence behavior of the population problem in this paper. More specifically, the population loss of matrix sensing and matrix completion is $\mathbb{E}f(Z)=c\|Z-X\|_F^2$ for some positive constant $c>0$, while that of the phase retrieval problem is $\mathbb{E}f(Z)=c\|Z-X\|_F^2+\frac{1}{2}\left(\|Z\|_F-\|X\|_F\right)^2$ with $c=1$ or $\frac{1}{2}$ (see Theorem \ref{thm: pr}). Thus we study the asymptotic behavior and the convergence rate of a sequence generated by randomly initialized\footnote{Randomly initialized means the initialization is drawn from the general random distribution (defined in Definition \ref{def: GRD}).} Riemannian gradient descent (the implementation of projected gradient descent defined in (\ref{eq: PGDalgorithm}), with details in Appendix \ref{sec: PGD}) minimizing
\begin{align}
        &\min_{Z\in\mathcal{M}_r}F_1(Z)=\frac{1}{2}\|Z-X\|_F^2, \label{eq: isoLS}\\
    \text{or }&\min_{Z\in\mathcal{M}_r} F_2(Z)= \frac{\theta}{2}(\|Z\|_F-\|X\|_F)^2+\|Z-X\|_F^2.\label{eq: weakerisoLS}
\end{align}

In other words, we analyze the isometry case $F_1(Z)=\frac{1}{2}\|Z-X\|_F^2$, and the weak isometry case $C_1\|Z-X\|_F^2\leq  F_2(Z) \leq C_2\|Z-X\|_F^2$, where $C_2>C_1>0$. Many low-rank matrix recovery problems belong to the former category, while the phase retrieval problem falls into the latter category (for more details see Section \ref{sec: StateMain}).

The main results of our global analysis framework are as follows.

\vspace{6pt}

\noindent{\bf{Theorem}} \textit{(Informal version of Theorem \ref{thm: rankrconverge}). With high probability no less than $1-\frac{1}{\text{poly}(n)}$, the sequence generated by a randomly initialized Riemannian gradient descent needs $\mathcal{O}(\log n+\log\frac{1}{\epsilon})$ iterations to reach an $\epsilon$-accurate solution of $X$ in minimizing (\ref{eq: isoLS}).}\footnote{$\Omega(N)$ means that there exist constants $C\geq c>0$ such that $c\cdot N\leq \Omega(N)\leq C\cdot N$, while $\mathcal{O}(N)$ means that there exist constant $C>0$ such that $\mathcal{O}(N)\leq C\cdot N$. }

\vspace{4pt}

\noindent {\bf{Theorem}} \textit{(Informal version of Theorem \ref{thm: pr}).
For a class of specific weak isometry problems (\ref{eq: weakerisoLS}), with high probability no less than $1-\frac{1}{\text{poly}(n)}$, the sequence generated by a random initialized Riemannian gradient descent needs $\mathcal{O}(\log n+\log\frac{1}{\epsilon})$ iterations to reach an $\epsilon$-accurate solution of $X$.}

\vspace{4pt}

The above results provide a partial explanation for the mechanism behind the nearly linear convergence rate of vanilla first-order methods. The  $\mathcal{O}(\log n)$ term in the number of iterations is mainly due to the fact that these problems have benign saddle points or saddle-like spurious points, and the sequence can escape these saddle points or saddle-like points in $\mathcal{O}(\log n)$ iterations. Under the setting of our problems, the vanilla Riemannian first-order scheme with random initialization converges to the second-order stationary points in a nearly linear rate essentially independent of the dimensionality of the problem with high probability.

\begin{figure}[htbp]
\centering
\begin{subfigure}[t]{0.30\textwidth}
\centering
\includegraphics[width=\linewidth]{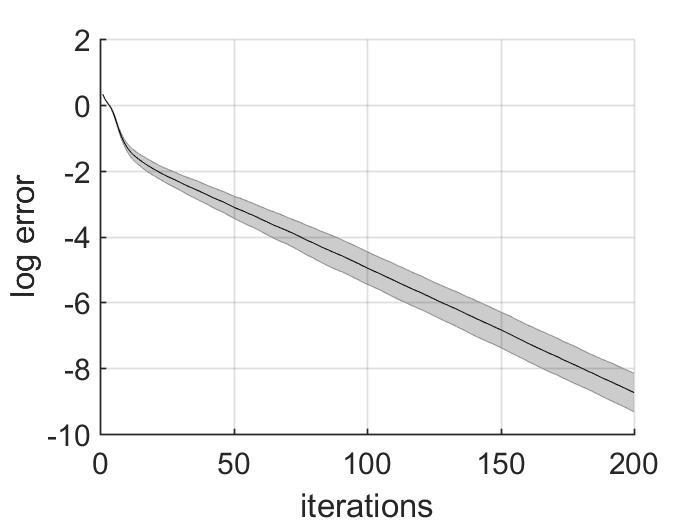}
\caption{\emph{Least squares function} with $F_1(Z)=\frac{1}{2}\|Z-X\|_F^2$, n=200, r=10.}\label{fig: LinearConverge_o}
\end{subfigure}
\hspace{0.03\textwidth}
\begin{subfigure}[t]{0.30\textwidth}
\centering
\includegraphics[width=\linewidth]{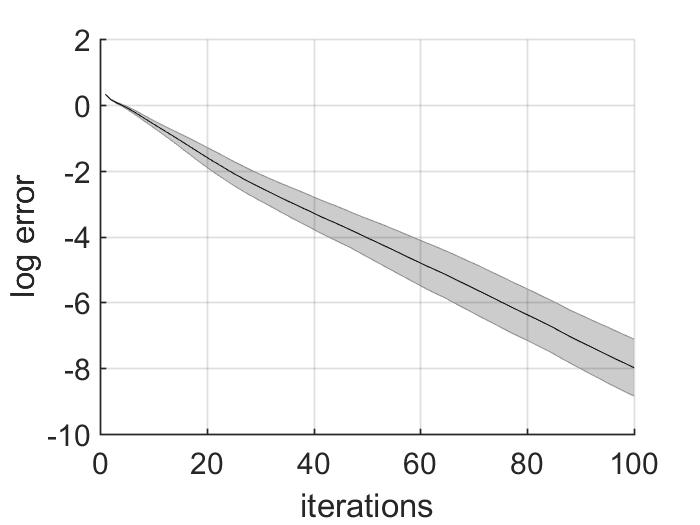}
\caption{\emph{Matrix sensing} with $f(Z)=\frac{1}{2}\|T(Z)-y\|_2^2$, n=100, r=5, m=2500.}\label{fig: LinearConverge_MS}
\end{subfigure}
\hspace{0.03\textwidth}
\begin{subfigure}[t]{0.30\textwidth}
\centering
\includegraphics[width=\linewidth]{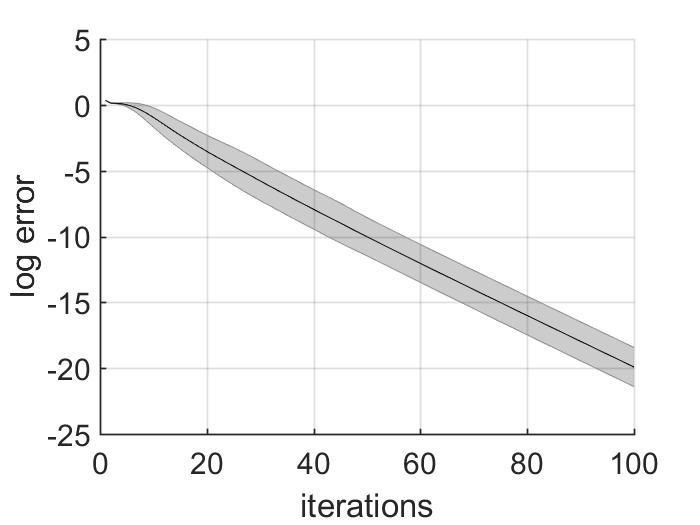}
\caption{\emph{Phase retrieval} with $f(Z)=\frac{1}{2}\|T(Z)-y\|_2^2$, n=100, m=1000.}\label{fig: LinearConverge_PR}
\end{subfigure}
\caption{
The PGD with random initialization. Each log-error band stands for the results from 100 independent experiments. }
\label{fig: IntroFigs}
\end{figure}

\smallskip
\noindent\textbf{Numerical illustration.} 
In Figure \ref{fig: IntroFigs}, we give some representative numerical results obtained using the Riemannian gradient descent (defined (\ref{eq: PGDalgorithm}) with details in Appendix \ref{sec: PGD}) with random initialization on the manifold to minimize the least squares loss functions of three problems.              
We observe nearly linear convergence in all three experiments. It is consistent with the theoretical results stated above. Note that the nearly linear convergence rate here is in sharp contrast with the sub-linear theoretical guarantees from previous works. Our goal is thus to bridge this gap between theory and practice.

\subsection{Comparison and challenges}

{\bf{Burer-Monteiro parametrization.}} Many low-rank matrix recovery problems are based on the Burer-Monteiro parameterization \cite{burer2003nonlinear,burer2005local,MaSen,MaCompletion,PhaseRetrieval}, which parameterizes a low-rank matrix $Z\in\F^{n\times n}$ by $UU^{\dag}$, with $U\in\F^{n\times r}$.\footnote{We assume that $X$ and $Z$ are SPSD (symmetric positive semidefinite). For a treatment of the asymmetric case, we refer to \cite{ge2017no}.}  It focuses on the least squares loss function as follows:
\begin{equation}\label{eq: euopt}
     \min_{U\in \mathbb{F}^{n\times r}}\frac{1}{2}\|T(UU^{\dag})-y\|_2^2.
\end{equation}
In contrast, we study a different formulation as follows
\begin{equation}\label{eq: maniopt2}
     \min_{Z\in \M_r}\frac{1}{2}\|T(Z)-y\|_2^2,
\end{equation}
and perform analysis directly in the manifold domain of $Z$ instead of the Euclidean domain of $U$. Problem (\ref{eq: maniopt2}) has a few advantages over Problem (\ref{eq: euopt}). Firstly, the Burer-Monteiro factorization could result in many more artificial critical points due to the reformulation. To see this, if $X=UU^\dag$ is a critical point, then $X=(U P)(U P)^\dag$ (the exponent $\dag$ denotes the Hermitian adjoint) with any unitary $P \in \F^{r\times r}$ is also a critical point. Such duplication of critical points can cause troubles both in the asymptotic behavior and the convergence rate \cite{paper1,du2017gradient}. Secondly, in many machine learning problems, the low-dimensional structure does not always translate into an appropriate factorization. Thus the global analysis on the low-rank matrix manifold provides a first step towards a broader extension to other applications in data science. Lastly, the objective function of (\ref{eq: maniopt}) is quadratic, while that of (\ref{eq: euopt}) is 4th-degree in terms of $U$. The former is more desirable in analysis.

\vspace{4pt}

\noindent{\bf{Rank-1 versus rank-r.}} There are plenty of works that explore the asymptotic landscape and exact convergence rate for rank-1 problems \cite{chen2019gradient,zhang2019sharp}. Our results differ from these previous works in that we provide a unified framework of analysis that applies to the general rank-r problem (Theorem \ref{thm: rankrconverge}), which is much more challenging than the rank-1 problem. To see the major technical challenge for general rank-r problems, note that while for the rank-1 problem, the core matrix is an $1\times 1$ matrix (a scalar), for general rank-r problem it becomes an $r\times r$ matrix. The closed form solutions of some quantities (e.g. the angles between the column spaces) are no longer available, which adds considerable difficulty to the convergence analysis.

\vspace{4pt}

\noindent{\bf{Convergence rate.}} Earlier landscape analysis on the low-rank matrix recovery \cite{MaSen,MaCompletion,PhaseRetrieval,li2019toward,robustPCA,ge2017no}, combined with the convergence guarantee for the nonconvex optimization \cite{TenDecomp,Jinchi2017}, indicates polynomial convergence towards the second-order stationary point. More recently, the authors in \cite{chen2019gradient} achieved nearly linear convergence for the rank-1 phase retrieval problem. 
 
We point out that the result of \cite{chen2019gradient} is consistent with our global analysis results (Section \ref{sec: StateMain}). In addition, we stress that the nearly linear convergence is common among other low-rank matrix recovery problems, at least from the manifold optimization perspective. Our work explores the following aspects: (1) whether the nearly optimal and fast convergence rate can be proved for the general rank-r matrix recovery; (2) how a weak isometry property affects the results; and (3) what is the common mechanism behind many different kinds of low-rank matrix recovery problems.


\subsection{Insights and discussions}
\noindent{\bf{The power of randomness.}} First-order schemes are widely used in large-scale computation due to their light computational cost. Under certain assumptions, first-order schemes are shown to have fast local convergence to the neighborhood of stationary points \cite{nesterov2013introductory,nesterov2006cubic}. A common problem with first-order schemes is that undesirable critical points including saddle points and local maxima could occur. Due to the lack of geometry information around critical points, first-order schemes with bad initialization could be trapped around the undesirable critical points instead of converging to the local minima. However, when augmented with randomness, the first-order schemes work well and have some provable guarantees. Below we discuss two major ways of incorporating randomness, namely the randomly perturbed first-order schemes and the randomly initialized first-order schemes.
\begin{itemize}
     \item \textit{Perturbed first-order schemes.} There are a few studies on the convergence of perturbed first-order schemes towards second-order stationary points both in the Euclidean and the Riemannian settings, see \cite{TenDecomp,Jinchi2017,du2017gradient,criscitiello2019efficiently,sun2019escaping}. These results show that general global convergence rate is polynomial and almost dimension-free. Whereas the intermittent perturbations help perturbed schemes escape the saddles better than non-perturbed first-order schemes in the worst case \cite{du2017gradient}, they also prevent a very accurate approximation of the ground truth without further (and sometimes complicated) modifications.
     
    \item \textit{Randomly initialized first-order schemes.} Though it has been proved that randomly initialized gradient descent asymptotically escapes saddles and only converges to the local minima  \cite{Jason2015,panageas2016gradient,Jason2017,paper1}, its convergence rate is much less clear.  In the worst case, when the initialization is close to the stable manifold of saddle points, the convergence towards the local minima slows down substantially. Indeed, the authors of the previous work \cite{du2017gradient} show that, in the worst case, the randomly initialized gradient descent can take exponential time to escape from the saddles. Despite such worst case scenario, the optimal efficiency of saddle escape behavior in a more general sense remains unclear. A recent answer to this question is given by the authors of \cite{chen2019gradient}, who show that for the rank-1 phase retrieval problem, gradient descent with random initialization has a nearly linear and almost dimension-free convergence rate, improving upon the previous polynomial convergence rate. This motivates us to study the mechanism behind the fast convergence rate and establish similar results for general rank-r matrix recovery problems.
\end{itemize}
We point out that an alternative way to avoid saddle points or local maxima is to use Hessian-based schemes and make use of second-order geometry information around the critical points. But the computational cost of such methods is expensive and its implementation can be complicated.

\vspace{6pt} 

\noindent{\bf{Population loss versus finite-sample.}}
Previous works \cite{chen2019gradient} have approached the heuristic treatment of the finite-sample loss function by splitting it into the population loss function plus the deviation from the population. We adopt a similar viewpoint on the Riemannian manifold.
 
For example, using the Riemannian gradient descent with random initialization to minimize 
\begin{align*}
    f(Z)=\frac{1}{2}\|T(Z)-T(X)\|_2^2, \quad Z\in\M_r,
\end{align*}
one can rewrite the empirical least-square loss function and its gradient into 
\begin{align*}
    f(Z)&= F(Z)+\frac{1}{m}\sum_{j=1}^m \theta_j, \qquad \theta_j = \langle A_j,Z-X\rangle^2-\mathbb{E}\left(\langle A_j,Z-X\rangle^2\right); \\
    \grad f(Z)&= \grad F(Z)+ \frac{1}{m}\sum_{j=1}^m \eta_j, \qquad \eta_j = \grad \theta_j.
\end{align*}
Here, $F(Z):=\mathbb{E}(f(Z))$, and $\theta_j$'s and $\eta_j$'s are i.i.d mean-zero random variables (deviation terms). By the central limit theorem, heuristically, with high probability we have
\begin{align*}
    \frac{1}{m}\sum_{j=1}^m\eta_j\lesssim \frac{\text{poly}(\log m)}{\sqrt{m}}.
\end{align*}
By requiring $m\gtrsim n\cdot \text{poly}(\log n)$, one can control the deviations $\sum_{j=1}^m\eta_j$. Then the minimization of the empirical loss function $f(Z)$ is close to that of the population loss function $F(Z)$. There are many methods to control the randomness. For example, the authors in \cite{chen2015solving,li2019toward} use truncation methods to obtain more delicate concentration properties. In this way, we can understand the mechanism behind many random sensing problems by analyzing the global convergence of the population dynamics.

\subsection{Organization of this paper}
The rest of the paper is organized as follows.
In Section \ref{sec: mainresults}, we present the main results of this paper, namely the framework of global analysis for three low-rank matrix recovery problems. We show the nearly linear convergence guarantee for the randomly initialized Riemannian gradient descent towards the ground truth, when minimizing different kinds of the population least squares loss functions.  
Specifically, Section \ref{sec: shortpre} gives brief preliminary information, Section \ref{sec: StateMain} states the main results, and Section \ref{sec: proofskech} highlights the main ideas of the proof. 
In Section \ref{sec: prooftech}, we lay out a full description of the proof strategy for the first main result in four parts. 
In particular, Section \ref{sec: KIS} describes the trajectory behavior, which serves as the key intermediate steps for proving the first main result.
Section \ref{sec: DLconverge} introduces the {\L}ojasiewicz convergence tool as a fundamental convergence guarantee.
Section \ref{sec: geometry_sr} is devoted to the geometry of the low-rank matrix manifold, where we classify all the critical points of the simple least squares function and study the spurious regions around spurious critical points. 
Some technical lemmas are given in Section \ref{sec: techlemma}. 
Section \ref{sec: prooftech2} contains the proofs of the second and third main results, which are similar but considerably simpler than that of the first main result.
Finally, in Section \ref{sec: Conclusion}, we make some concluding remarks. 
In Appendix \ref{sec: pre}, we present the manifold setting and the light-computational optimization technique. 
All technical proofs are deferred to Appendix \ref{sec: Proofs}, and some auxiliary lemmas are provided in Appendix \ref{sec: auxLM}.
\section{Main results}\label{sec: mainresults}
In this section, we introduce the main results of this paper. The optimization problems we are looking at are (\ref{eq: isoLS}) and (\ref{eq: weakerisoLS}), namely we minimize either $F_1(Z)=\frac{1}{2}\|Z-X\|_F^2$ or $F_2(Z)= \frac{\theta}{2}(\|Z\|_F-\|X\|_F)^2+\|Z-X\|_F^2$ on the low-rank matrix manifold $\M_r$. The analysis is based on the case of symmetric positive semi-definite (SPSD), and one may refer to \cite{ge2017no} for the treatment of asymmetric case.

\subsection{Preliminaries}\label{sec: shortpre}

We first introduce the necessary notations, optimization techniques and assumptions for the statement of our main results.

\vspace{8pt}

\noindent{\bf{Notations.}}
Denote $\M_s$ as the fixed rank manifold $\{Z\in\F^{n_1\times n_2}: \text{rank}(Z)=s\}$, and $\overline{\M_s}$ as its closure. In the symmetric (Hermitian) case, denote $n=n_1=n_2$. Throughout the paper, $\F=\R$ or $\F=\C$. Let  $(\cdot)^\dag$ denote the transpose (adjoint), i.e. $(\cdot)^\dag=(\cdot)^\top$ when $\F=\R$; and $(\cdot)^\dag=(\cdot)^*$ when $\F=\C$. 
The matrix $X$ always denotes the ground truth matrix with $\text{rank}(X)=r$, while $Z^*$ can be any fixed point (or accumulating point) of an algorithm. 
For integer $s>0$, denote $[s]=\{1,2,...,s\}$. For any $j\in[r]$, let $\setminus j:=[r]\setminus \{j\}$.
Denote $X=UDU^{\dag}$. For any $j\in[r]$, let $d_j:=D(j,j)$, and $U_j:=U(:,j)$. Let $\sigma_j(\cdot)$ denote the $j$-th largest singular value or eigenvalue of a matrix.
The $d_j$'s and $\sigma_j$'s are in the descending order unless otherwise specified.
Unless otherwise specified, the vector norm we use is $\|\cdot\|_2$ and the matrix norm is $\|\cdot\|_F$. Unless otherwise specified, we use $0<c,\,C<\infty$ to denote any absolute constant independent of $n$ in our statement that may vary in different contexts. The symbol $\Omega(N)$ means that there exist constants $C\geq c>0$ such that $c\cdot N\leq \Omega(N)\leq C\cdot N$, and $\mathcal{O}(N)$ means that there exist a constant $C>0$ such that $\mathcal{O}(N)\leq C\cdot N$. In this paper, we focus on the large $n$ regime, and other quantities including $r$ and $\sigma_r(X)=d_r$ will be treated as constants and will be ignored in the $\mathcal{O}(\cdot)$ and $\Omega(\cdot)$. The symbol $\text{poly}(n)$ stands for a nonnegative quantity upper bounded by $C\cdot n^k$ for some $C>0$ and $k\in\mathbb{N}_+$.
\vspace{8pt}

\noindent{\bf{The optimization technique.}} We use a Riemannian gradient descent method called the Projected Gradient Descent (the PGD):

\begin{equation}\label{eq: PGDalgorithm}
    Z_{k+1}=\mathcal{R}\left(Z_k - \alpha_k P_{T_{Z_k}}(\grad f(Z_k))\right).
\end{equation}
Here $P_{T_{Z_k}}$ is the projection onto the tangent space of $\M$ at point $Z_k$, $\alpha_k$ is the $k$-th stepsize, and $\mathcal{R}: T_{Z} \rightarrow \M$ is a retraction operator. This can be viewed as the vanilla first-order scheme on the Riemannian manifold. A more detailed description of the method and the respective operations are deferred to Appendix \ref{sec: pre}. 

\vspace{8pt}

\noindent{\bf{Assumptions.}} The following assumptions are necessary for all of the main results.

\vspace{4pt}

\noindent{Assumption 1}:  Assume $\alpha>0$ is a  small constant such that the discretized system can be well approximated by the continuous system. In other words, let $A(t)$ denote a continuous system of interest and $\{A_k\}$ be its time discretization, then we assume that $A_{k+1}=A_k+\alpha \dot{A}_k+o(\alpha \dot{A}_k)$.

\vspace{4pt}

\noindent{Assumption 2: Assume that the singular values of $Z$ are always simple and do not cross one another along the whole gradient flow or gradient descent trajectory. Moreover, the smallest singular value gap is lower bounded  along the whole trajectory, i.e. $\min_{i\neq j}|\sigma_i(Z) -\sigma_j(Z)|\ge c_g\sigma_1(Z)$.}

\vspace{4pt}

\begin{remark}
    Gradient flow of $Z$ with non-crossing singular values is common in practice. In fact, it is \emph{generic}\footnote{A property of a topological space is called \emph{generic} if it holds for a subset of the space which is of the second Baire category.} as is proved in \cite{dieci1999smooth}. However, gradient flows with crossing singular values have also been observed in some experiments. Assumption 2 is mainly for the purpose of simplifying the presentation of our technical analysis, as crossing singular values would introduce additional difficulties. These additional difficulties could be overcome by considering subspaces of singular vectors as a whole. This is left for future work. 
\end{remark}

\vspace{4pt}

\begin{figure}[ht]
     \centering
     \begin{subfigure}[b]{0.42\textwidth}
         \centering
         \includegraphics[width=\textwidth]{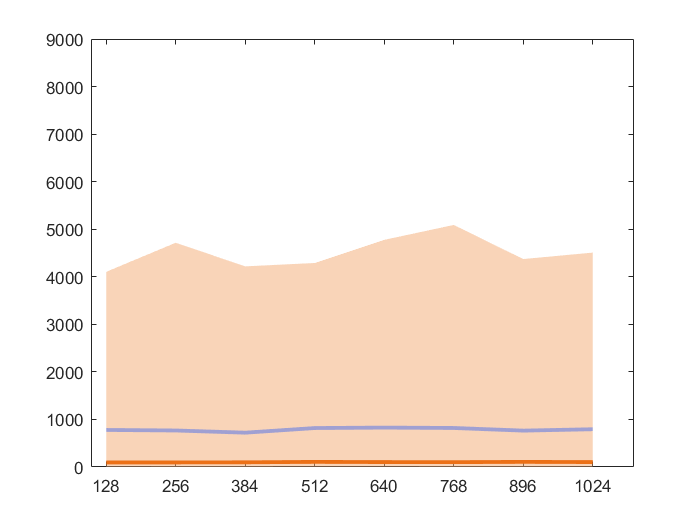}
         \caption{Bound for $L$}
     \end{subfigure}
     \hspace{0.1\textwidth}
     \begin{subfigure}[b]{0.42\textwidth}
         \centering
         \includegraphics[width=\textwidth]{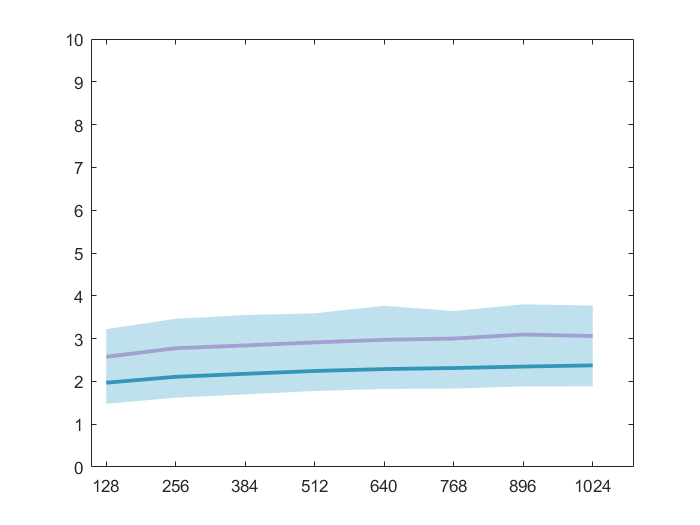}
         \caption{Bound for $C_u$}
     \end{subfigure}
     \caption{\small Results of 3000 independent numerical experiments with fixed rank=5, and dimension $n=128,256,384,512,640,768,896,1024$. Shaded regions represent the quantile $[0,0.98]$. The purple line stands for the 95\% quantile, while the orange and blue line stands for the median.
     \label{fig: assumption3}
     }
\end{figure}

\noindent{Assumption 3: Let $Z=V_z\Sigma_zV_z^{\dag}$ and $X=UDU^{\dag}$ be the eigenvalue decompositions of $Z$ and $X$. Let $R =  V_z^\dag U$. We assume that, along the whole trajectory,
\begin{enumerate}[label=(\arabic*)]
    \item There exists an absolute constant $L>0$ such that for any $i\in[r]$, we have $|R(i,j)|\leq L |R(i,i)|$ and $|R(i,j)|\leq L |R(j,j)|$ for $j\in[r]\setminus \{i\}$ at each iteration;
    \item The ratio $\frac{(RDR^{\dag})(i,i)}{\sigma_i(Z)}$ is upper bounded, i.e. there exists $C_u>0$ such that $\frac{(RDR^{\dag})(i,i)}{\sigma_i(Z)}<C_u$.
\end{enumerate} 

\begin{remark}
We provide some numerical experiments in Figure \ref{fig: assumption3} to demonstrate that Assumption 3 is a reasonable assumption in that we can find such constants $L$ and $C_u$ as upper bounds with high probability. The figures validate the existence of both $L$ and $C_u$. 
\end{remark}
}

\subsection{Statement of main results}
\label{sec: StateMain}
We are now ready to state the three main results of this paper. 

\vspace{8pt}

\noindent{\bf{Main result 1: Global convergence for least squares on $\M_r$.}}
We will prove that the randomly initialized Riemannian gradient descent (the PGD) for $F_1(Z) = \frac{1}{2}\|Z-X\|_F^2$ with high probability escapes the spurious critical points $\mathcal{S}_{Z^*}$ (to be defined in Section \ref{sec: geometry_sr}) and converges to the global minimum $X$. Moreover, with high probability the convergence rate is nearly linear and almost dimension-free, i.e., for  $k\gtrsim \mathcal{O}(\log n)$, we have $\|Z_k-X\|_F\leq e^{-ck}\|Z_0-X\|_F$, with the constant $c>0$ uniformly bounded from below. Specifically, we have the following theorem.

\begin{theorem}
\label{thm: rankrconverge}
    Let $F_1(Z) = \frac{1}{2}\|Z-X\|_F^2$, and $X=UDU^{\dag}$ has distinct eigenvalues.
    Let $\{Z_k\}_{k=0}^\infty$ be the sequence generated by the Riemannian gradient descent (the PGD) initialized at $Z_0$ which is drawn from the general random distribution (as defined in Definition \ref{def: GRD}). Under Assumption 1, 2 and 3, there exists constant stepsize $\alpha>0$ small enough such that with high probability no less than $1-\frac{1}{\text{poly}(n)}$, we have the following convergence results:
    \begin{enumerate}[label=\arabic*)]
        \item  The initial matrix $Z_0$ falls into the branch such that $\lim_{k\rightarrow\infty}Z_k=X$. Further, there exists $\mathcal{T}=\mathcal{O}(\log n)$ such that:
        \begin{align*}
            \|Z_k-X\|_F^2 \le e^{-ck},\text{ for all } k\geq \mathcal{T}.
        \end{align*}
       The positive constant $c$ depends on $d_r$ and $\alpha$. In other words, it takes no more than $\mathcal{O}(\log \frac{1}{\epsilon}+\log n)$ iterations to get an $\epsilon$-accurate solution (i.e. to achieve $\|Z-X\|\leq \epsilon \|X\|_F$) via the randomly initialized Riemannian gradient descent (the PGD).
        \item With probability 0, $Z_0$ falls into the other branches such that $\lim_{k\rightarrow\infty}Z_k=Z^*\in\mathcal{S}_{Z^*}$ (defined in Lemma \ref{lem: spurious_fp}).
    \end{enumerate}
\end{theorem}

We will prove Theorem \ref{thm: rankrconverge} by using Theorem \ref{thm: stage1}, Theorem \ref{thm: stage2} and Theorem \ref{thm: stage3}, and will provide a detailed proof strategy in Section \ref{sec: prooftech}.

\vspace{6pt}

\noindent{\textit{Stationary points.}} For $F_1(Z)=\frac{1}{2}\|Z-X\|_F^2$, the stationary points of the PGD are either the ground truth $X$, or 
$$\mathcal{S}_{Z^*} := \left\{Z^*: \, Z^* = U_1D_1 V_1^{\dag}, \text{ where } U = \left(U_1, U_2\right), \, D = \text{diag } \{D_1, D_2\} \text{ and }  V = \left(V_1, V_2\right) ,Z^*\neq X\right\}.$$
See also Lemma \ref{lem: spurious_fp} for more details of this result. If $X$ has distinct eigenvalues, then $|\mathcal{S}_{Z^*}|=2^r-1$. 
In particular, every point in $\mathcal{S}_{Z^*}$ has a property similar to that of a saddle point (we call it a saddle-like property), except that the Hessian has a $-\infty$ direction, which indicates that the curvature is singular at such point. 
The neighborhood of $\mathcal{S}_{Z^*}$ is crucial because the Riemannian gradient becomes degenerate. 

\medskip
\noindent{\textit{Attraction by spurious stationary points.}} The fundamental convergence tool to be presented in Theorem \ref{thm: LD2converge} ensures linear convergence towards $X$ when minimizing $F_1(Z)$ via the Riemannian gradient descent (the PGD) in most part of the manifold. However, when the the sequence gets close to $\mathcal{S}_{Z^*}$, in the local region $\mathcal{B}(\mathcal{S}_{Z^*},\delta)$, the convergence slows down and the sequence is attracted to $\mathcal{S}_{Z^*}$. 
Although the worst initialization leads to non-escape from $\mathcal{S}_{Z^*}$ (or, say $\infty$ number of iterations to escape), we estimate the upper bound on the number of iterations needed to escape $\mathcal{B}(\mathcal{S}_{Z^*},\delta)$ in the sense of with high probability.

\medskip
\noindent{\textit{Estimation of the iterations trapped by $\mathcal{S}_{Z^*}$.}} From Lemma \ref{lem: uniformST}, the ``angle'' (the product of the two column vector matrices) between the randomly initialized column space and the ground truth column space is of order $\Omega(\frac{1}{n})$.  If the angle remains $\Omega(\frac{1}{n})$ when the sequence enters the spurious region  $\mathcal{B}(\mathcal{S}_{Z^*},\delta)$ (defined in Lemma \ref{lem: balls}) and it grows exponentially fast, then it takes $\mathcal{O}(\log n)$ iterations for the angle to become $\Omega(1)$. By Lemma \ref{lem: ness}, this indicates successful escape from the spurious regions. To further understand why it takes $\mathcal{O}(\log n)$ iterations to escape the spurious regions, we use a toy example (Example \ref{eg: euclieg}) to show that for a general strict saddle, exponential escape behavior is also present. However, the $Z^*$ differs from such a general strict saddle in that the Hessian of $Z^*$ has $-\infty$ directions. Still, we can bound the number of iterations needed to escape from $\mathcal{B}(\mathcal{S}_{Z^*},\delta)$ by throwing away a small probability measure of points. Such analysis is based on Lemma \ref{lem: ness}, Lemma \ref{lem: uniformST} and Lemma \ref{lem: Rdynamics}.

\vspace{10pt}

\noindent{\bf{Main result 2: Global convergence for  least squares on $\M_1$.}}
The second main result can be regarded as a special case of Main result 1 restricted to $r=1$. However, the analysis is much easier to follow. We list this as one of the main results because the trajectory behavior of this case is very similar to that of phase retrieval and its generalization in the third main result. When $r=1$, by Lemma \ref{lem: spurious_fp}, we have  $\mathcal{S}_{Z^*}=\{0\}$. In addition, we can directly write out the closed form gradient descent or gradient flow. The result is stated in the following theorem.
\begin{theorem}
\label{thm: rank1thm}
Assume $X$ satisfies $\text{rank}(X)=1$, $\|X\|_F=1$. We consider $F_1(Z)=\frac{1}{2}\|Z-X\|_F^2$ with $Z\in\mathcal{M}_1$. Let $\{Z_k\}$ be the sequence generated by the Riemannian gradient descent (the PGD) initialized at $Z_0$ which is drawn from the general random distribution. Denote $Z=zz^{\dag}$ and $X=xx^{\dag}$, $h=\|Z\|_F$ and $\rho=\frac{\langle X,Z\rangle}{\|X\|_F\|Z\|_F}$. Then, we have
\begin{enumerate}[label=\arabic*)]
\item 
The continuous evolution dynamics of the gradient flow can be described by the following ODE system:
\begin{align*}
    \frac{d}{dt}h&=-h+\rho,\\
    \frac{d}{dt}\rho&=2\frac{\rho}{h}(1-\rho).
\end{align*}
Consequently, with high probability no less than $1-\frac{1}{\poly(n)}$, the Riemannian gradient flow only converges to $Z^*=X$, and it takes $\mathcal{O}(\log n+\log\frac{1}{\epsilon})$ time to generate an $\epsilon$-accuracy solution, i.e. to achieve $\|Z-X\|_F\leq\epsilon\|X\|_F$.

\item In addition, the discrete evolution dynamics can be described by the following discrete system:
\begin{align*}
    h_{k+1}&=h_k+\alpha(-h_k+\rho_k)+\mathcal{O}(\alpha^2),\\
    \rho_{k+1}&=\rho_k+2\alpha\frac{\rho_k}{h_k}(1-\rho_k)+\mathcal{O}(\alpha^2).
\end{align*}
Under Assumption 1, there exists $\alpha>0$ small enough such that with high probability no less than $1-\frac{1}{\text{poly}(n)}$, the Riemannian gradient descent (the PGD) only converges to the equilibrium with $\rho^*=h^*=1$, meaning $\lim_{k \to \infty} Z_{k}=X$. Moreover, it takes $\mathcal{O}(\log \frac{1}{\epsilon}+\log n)$ iterations to generate an $\epsilon$-accurate solution, i.e. to achieve $\|Z-X\|_F\leq\epsilon\|X\|_F$.
\end{enumerate}
\end{theorem}

Theorem \ref{thm: rank1thm} is in preparation for the next main result on the global convergence for the special case of phase retrieval. As we will see below, the population loss function of phase retrieval and its generalization differs from $F_1$ in that it only satisfies a weak isometry property.
A detailed proof of Theorem \ref{thm: rank1thm} and its connection with Theorem \ref{thm: pr} can be found in Section \ref{sec: prooftech2}. The basic idea of the proof is similar to that of Theorem \ref{thm: rankrconverge} but much simpler.

\vspace{10pt}

\noindent{\bf{Main result 3: Global convergence for the population phase retrieval problem.}}
Isometry properties weaker than the RIP (Restricted Isometry Property) are also common in various real-world applications. An example is the phase retrieval problem, whose loss function is given below:
\begin{equation}
\label{eq: prloss}
    f(Z) =\frac{1}{2}\|T(Z)-y\|_2^2\nonumber =\frac{1}{2m}\sum_{j=1}^m \langle A_j,Z-X\rangle^2.
\end{equation}
Here, $X=xx^{\dag}$ is the ground truth, $y = T(X)$, $Z=zz^{\dag}$, and $A_j=a_ja_j^{\dag}$, where $a_j$'s are i.i.d drawn from $\mathcal{N}(0,I_n)$ or $\frac{1}{\sqrt{2}}\left(\mathcal{N}(0,I_n)+i\cdot \mathcal{N}(0,I_n)\right)$. 
To simplify the analysis, we only establish the result for the population loss function here. We focus on studying the problem on the Riemannian manifold and revealing its connection to the rank-1 isometry case (Theorem \ref{thm: rank1thm}). Our proof complements that of \cite{chen2019gradient}, which establishes a complete proof for random measurements from a different viewpoint. 
\begin{theorem}\label{thm: pr}
For the Gaussian phase retrieval problem (\ref{eq: prloss}), we have the following results
\begin{enumerate}[label=\arabic*)]
    \item Let  $F_2 (Z) := \mathbb{E} f(Z)$ be the population loss of (\ref{eq: prloss}). Then, we have $F_2(Z)=\frac{\theta}{2}(\|Z\|_F-\|X\|_F)^2+c\cdot\|Z-X\|_F^2$, where $\theta=1$, and $c=1$ when $\F=\R$, or $c=\frac{1}{2}$ when $\F=\C$.
 
\item Consider 
$F_2(Z)=\|T(Z)-T(X)\|_2^2=\frac{\theta}{2}(\|Z\|_F-\|X\|_F)^2+\|Z-X\|_F^2$, with $0<\theta<\Omega(1)$ and $T$ satisfies $C_1\|Z'-Z\|_F\leq \|T(Z')-T(Z)\|\leq C_2\|Z'-Z\|_F$ for $Z$, $Z'\in\M_1$ with some $C_2\geq C_1>0$. Denote $Z=zz^{\dag}$, $X=xx^{\dag}$, $h=\|Z\|_F$ and $\rho=\frac{\langle X,Z
\rangle}{\|X\|_F\|Z\|_F}$. Then, the continuous evolution dynamics of the gradient flow can be described by the following ODE system:
\begin{align*}
    \frac{d}{dt}h&=\theta-(2+\theta)h+2\rho,\\
    \frac{d}{dt}\rho&=\frac{4\rho}{h}(1-\rho).
\end{align*}

Consequently, with high probability no less than $1-\frac{1}{\poly(n)}$, the Riemannian gradient flow only converges to $Z^*=X$, and it takes $\mathcal{O}(\log n+\log\frac{1}{\epsilon})$ time to generate an $\epsilon$-accuracy solution, i.e. to achieve $\|Z-X\|_F\leq\epsilon\|X\|_F$.

\item Let $\{Z_k\}$ be the sequence generated by the Riemannian gradient descent (the PGD) initialized at $Z_0$ which is drawn from the general random distribution. Then the discrete evolution dynamics can be described by the following discrete system:
\begin{align*}
    h_{k+1}&=h_k+\alpha(\theta-(2+\theta)h_k+2\rho_k)+\mathcal{O}(\alpha^2),\\
    \rho_{k+1}&=\rho_k+\alpha\frac{4\rho_k}{h_k}(1-\rho_k)+\mathcal{O}(\alpha^2).
\end{align*}
Under Assumption 1, there exists $\alpha>0$ small enough such that with high probability no less than $1-\frac{1}{\text{poly}(n)}$, the Riemannian gradient descent (the PGD) converges to the equilibrium with $\rho^*=h^*=1$ only, meaning $\lim_{k \to \infty} Z_{k}=X$. Moreover, it takes $\mathcal{O}(\log \frac{1}{\epsilon}+\log n)$ iterations to generate an $\epsilon$-accurate solution, i.e. $\|Z-X\|_F\leq\epsilon\|X\|_F$.
\end{enumerate}
\end{theorem}

The proof of Theorem \ref{thm: pr} is built upon Theorem \ref{thm: rank1thm}, as $F_2$ satisfies a weaker isometry property than that of $F_1$. The detailed proof is given in Section \ref{sec: prooftech2}.

\subsection{Sketch of proof}\label{sec: proofskech}
Here we highlight a few high level ideas of our proofs for the main theorems stated in the previous subsection. For a more comprehensive and detailed presentation of the proof strategy and intermediate results, the reader may refer to Section \ref{sec: prooftech} and Section \ref{sec: prooftech2}.

\vspace{6pt}

\begin{figure}[ht]
        \centering
        \includegraphics[width=0.4\textwidth]{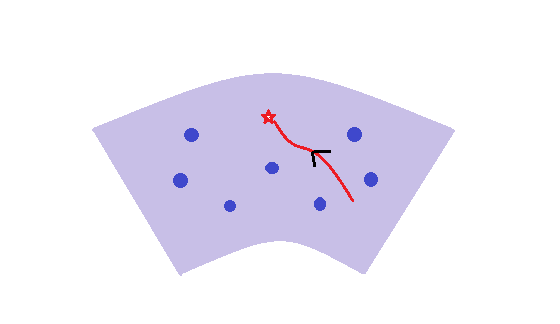}
        \caption{Illustration of the trajectory of the PGD on $\M_r$}
        \label{fig: majarea-copy}
\end{figure}

\noindent\textbf{Fundamental convergence guarantee by the \L ojasiewicz inequality.} The \L ojasiewicz inequality \cite{convergence2005absil,attouch2010proximal,bolte2010characterizations,Reinhold} has long been studied as a fundamental tool for convergence analysis. It is especially useful in proving the linear convergence rate of first-order optimization methods. In Section \ref{sec: DLconverge}, we derive a version of this tool tailored for the PGD method, stated in Theorem \ref{thm: LD2converge}. Using this theorem, the task of checking the convergence rate of the PGD for a specific problem is reduced to checking Conditions (\ref{eq: D}) and (\ref{eq: L}) for the objective function. We then observe that for Problems (\ref{eq: isoLS}) and (\ref{eq: weakerisoLS}), these conditions are satisfied except for some small regions on the manifold. Such regions are later dubbed \emph{spurious regions}, see e.g. the blue balls in Figure \ref{fig: majarea-copy}. This leads to the next important result on the geometry of the low-rank matrix manifold. 
\vspace{6pt}

\noindent\textbf{Geometry of spurious regions over $\M_r$.} The low-rank matrix manifold $\M_r$ is a non-closed set. This fact poses special challenges to our analysis. We find out that for the simple least squares loss function $F_1(Z) = \frac{1}{2}\|Z-X\|_F^2$, apart from the ground truth solution $Z=X$, there are a few \emph{spurious critical points} in $\overline{\M_r}\backslash\M_r$, which we denote as $\mathcal{S}_{Z^*}$. In the local region of these spurious critical points, there are certain \emph{spurious regions} where Condition (\ref{eq: L}) could fail. In other words, outside the spurious regions, linear convergence rate is guaranteed; while inside the spurious regions, the convergence rate slows down. It then becomes important to characterize how the PGD may escape these regions and converge to the ground truth. A full description of the spurious critical points and spurious regions, along with examples and illustrations, is given in Section \ref{sec: geometry_sr}.

\vspace{6pt}

\noindent\textbf{Three-stage description of the trajectory behavior.} It now remains to study how the trajectory of the randomly initialized gradient descent sequence escapes the spurious regions and converges to the ground truth on the manifold. In Section \ref{sec: KIS}, we divide the whole trajectory into three stages. In the worst case, the sequence is dragged towards $Z^* = 0$ in the first stage, then escapes $Z^* = 0$ and approaches some other  $Z^* \in \mathcal{S}_{Z^*}\setminus \{0\}$ in the second stage, and finally escapes such $Z^*$ in the third stage. We show that by throwing out a small probability measure, the total number of iterations needed to reach the $\epsilon$-neighborhood of $X$ is bounded by $\mathcal{O}(\log(\frac{1}{\epsilon})+\log n)$, as stated in the main theorems. This worst case is however rarely observed in numerical experiments; usually, the sequence generated by the PGD simply avoids all spurious critical points and converges straight to the ground truth.  Proofs of the three-stage behavior are stated in Theorem \ref{thm: stage1}, Theorem \ref{thm: stage2}, Theorem \ref{thm: stage3}, and rely on a series of technical lemmas detailed in Section \ref{sec: techlemma}.

\section{Riemannian global analysis and proof of Theorem \ref{thm: rankrconverge}}\label{sec: prooftech}
This section is primarily devoted to the proof of Theorem \ref{thm: rankrconverge}. This theorem is the central result of this paper and fully describes the trajectory behavior of the PGD for the least squares loss function on the rank-$r$ matrix manifold $\M_r$. The proofs of Theorem \ref{thm: rank1thm} and \ref{thm: pr} in Section \ref{sec: prooftech2} mainly follow that of Theorem \ref{thm: rankrconverge}, which serves as a general framework for the global analysis. Their proofs also depend on the technical results in Sections \ref{sec: DLconverge} and \ref{sec: techlemma} but differ in some specific aspects.

We introduce some more notations to be used here. For any $Z\in\M_r$, $B(Z,\delta):=\{Z':\|Z-Z\|_F\leq \delta, \, Z'\in \M_r\}$ is the neighborhood, while $\mathcal{B}(\cdot,\delta)$ defined in Lemma \ref{lem: balls} is the ``spurious region''.

\subsection{Key intermediate steps for Theorem \ref{thm: rankrconverge}}\label{sec: KIS}

To prove Theorem \ref{thm: rankrconverge}, we divide the trajectory of the PGD on the manifold into three stages as follows.

\vspace{6pt}

\noindent{\bf{Stage 1:}} The iterative sequence starts from a random initialization point. For iteration $k\lesssim \frac{1}{\alpha\delta^2}\log\frac{1}{\delta}$, with high probability exceeding $1-\frac{1}{\text{poly}(n)}$, the sequence either enters the $\delta$-local region of $X$ or reaches the local region of $Z^*=0$.  \\
{\bf{Stage 2:}} If the sequence enters the $\delta$-local region of $X$, it will converge linearly to the target $X$, so it further takes $\mathcal{O}(\log\frac{1}{\epsilon})$ iterations to generate an $\epsilon$-accurate solution. On the other hand, if the sequence reaches the $\delta$-neighborhood of $Z^*=0$, it takes $\mathcal{O}(\log n)$ iterations to escape the local region, and enters the stage 3.\\
{\bf{Stage 3:}} The sequence either  enters the $\delta$-local region of $X$ without getting close to any $Z^*\in\mathcal{S}_{Z^*}\setminus \{0\}$, or, reaches a $\delta$-spurious region of other $Z^*\in\mathcal{S}_{Z^*}\setminus \{0\}$. If it enters the $\delta$-local region of $X$, it takes $\mathcal{O}(\log\frac{1}{\epsilon})$ iterations to generate an $\epsilon$-accurate solution. On the other hand, if it reaches a $\delta$-spurious region of some $Z^*\in \mathcal{S}_{Z^*}\setminus \{0\}$, then with high probability after $\mathcal{O}(\log n)$ iterations the sequence will escape the $\delta$-spurious region and take additional
    $\mathcal{O}(\log\frac{1}{\epsilon})$ iterations to generate an $\epsilon$-accurate solution.

The following three theorems state the above three stages in mathematical terms respectively. Their proofs are given in Section \ref{sec: proof3stage}.

\begin{theorem}\label{thm: stage1}
Under the setting of Theorem \ref{thm: rankrconverge},
there exists stepsize $\alpha>0$ and a small enough constant $\delta>0$ such that with high probability exceeding $1-\frac{1}{\text{poly}(n)}$, for all $k\leq K_0=\mathcal{O}(1)$, we have $\phi^k(Z_0)\in B(X,\delta)\cup\mathcal{B}(0,\delta)\setminus \cup_{Z^*\in\mathcal{S}_{Z^*}\setminus \{0\}}\mathcal{B}(Z^*,\delta)$.
Here, the constant in the fail probability depends on $\alpha$, $\delta$ and $K_0$.
\end{theorem}

\begin{theorem}\label{thm: stage2}
Under the setting of Theorem \ref{thm: rankrconverge}, suppose $K_0$ is a positive integer such that $\left(\cup_{i=1}^{K_0-1}Z_i\right)\cap \mathcal{B}(0,\delta)=\emptyset$ and $Z_{K_0}\in\mathcal{B}(0,
\delta)$. Then, there exists stepsize $\alpha>0$, a small enough constant $\delta>0$ and an absolute constant $C_1=\mathcal{O}(1)>0$ such that with high probability exceeding $1-\frac{1}{\text{poly}(n)}$, we have $\cup_{i=K_1}^\infty Z_i\cap \mathcal{B}(0,\delta)=\emptyset$. Here, $K_1=K_0+C_1\log n$.
\end{theorem}

\begin{theorem}\label{thm: stage3}
Under the setting of Theorem \ref{thm: rankrconverge}, for any $Z^*\in\mathcal{S}_{Z^*}\setminus\{0\}$, suppose there exists a positive integer $K_1$ such that $\{\cup_{i=1}^{K_1-1}Z_i\}\cap \mathcal{B}(\mathcal{S}_{Z^*}\setminus \{0\},\delta)=\emptyset$ and $Z_{K_1}\in\mathcal{B}(Z^*,
\delta)$. Then, there exists stepsize $\alpha>0$, a small enough constant $\delta>0$ and $C_2=\mathcal{O}(1)$ such that with high probability exceeding $1-\frac{1}{\text{poly}(n)}$, we have $\cup_{i=K_2}^\infty Z_i\cap \mathcal{B}(Z^*,\delta)=\emptyset$. Here, $K_2=K_1+C_2\log n$.
\end{theorem}

\tikzstyle{startstop} = [rectangle, rounded corners, minimum width = 2cm, minimum height=1.25cm,text centered, draw = black, fill = red!20]
\tikzstyle{process} = [rectangle, rounded corners, minimum width=4cm, minimum height=1.5cm, text centered, draw=black, fill = green!15]
\tikzstyle{decision} = [rectangle, rounded corners, aspect = 3, text centered, draw=black, fill = blue!15]
\tikzstyle{arrow} = [->,>=stealth]
\begin{tikzpicture}[node distance=2cm]
\node (target) [startstop]
{Proof of Theorem \ref{thm: rankrconverge}};
\node (trajectory) [process, below of = target]{\shortstack{Trajectory: three main intermediate stages, Theorem \ref{thm: stage1}, \ref{thm: stage2}, \ref{thm: stage3}\\(Proofs are given in Section \ref{sec: proof3stage})}};
\node (tools2) [decision, below of = trajectory]{\shortstack{Spurious regions \\(Section \ref{sec: geometry_sr})}};
\node (tools1) [decision, left of = tools2, xshift = -3cm]{\shortstack{\L ojasiewicz convergence\\ tool (Section \ref{sec: DLconverge})}};
\node (tools3) [decision, right of = tools2, xshift = 3.5cm]{\shortstack{Technical lemmas\\ (Section \ref{sec: techlemma})}};
\draw [arrow] (target) -- (trajectory);
\draw [arrow] (trajectory) -- (tools2);
\draw [arrow] (trajectory) -- (tools1);
\draw [arrow] (trajectory) -- (tools3);
\end{tikzpicture}

\begin{proof}[\textbf{Proof of Theorem \ref{thm: rankrconverge}}]
Let the stepsize $\alpha>0$ and $0<\delta<\frac{d_r}{2}$ be small enough constants that meet the requirements in Theorem \ref{thm: stage1}, Theorem \ref{thm: stage2} and Theorem \ref{thm: stage3}. Then, by Lemma \ref{lem: P_Tnorm}, we have linear convergence in the local neighborhood of $X$. In addition, by Lemma \ref{lem: balls} and Lemma \ref{lem: LS}, such linear convergence is true in the majority of the manifold except for the spurious regions. By Lemma \ref{lem: F12singlestep}, if $Z_k \in \mathcal{M}_r\setminus \mathcal{B}(\mathcal{S}_{Z^*},\delta)$, then $\|Z_{k+1}-X\|_F \leq e^{-c}\|Z_k-X\|_F$. Here, $c=-\log(1-\Omega(\alpha\delta^2))$. If for all $k\leq K_0=\mathcal{O}(\frac{1}{\alpha\delta^2}\log \frac{1}{\delta})$, $Z_k\in\mathcal{M}_r\setminus \mathcal{B}(\mathcal{S}_{Z^*},\delta)$, then $Z_{K_0}$ enters the neighborhood of the ground truth $X$, i.e. $Z_{K_0}\in B(X,\delta)$. By Lemma \ref{lem: P_Tnorm}, we know that it takes additional $\mathcal{O}(\log \frac{1}{\epsilon})$ iterations to generate an $\epsilon$-accurate solution. On the other hand, if there exists $k\leq K_0=\mathcal{O}(\frac{1}{\alpha\delta^2}\log \frac{1}{\delta})$ such that $Z_k$ enters the spurious region of $Z^*$, i.e.  $Z_k\in \mathcal{B}(\mathcal{S}_{Z^*},\delta)$, then we know from Theorem \ref{thm: stage1} that with high probability exceeding $1-\frac{1}{\text{poly}(n)}$, we have $Z^*=0$. By Theorem \ref{thm: stage2}, it then takes $\mathcal{O}(\log n)$ iterations to escape $\mathcal{B}(0,\delta)$, with the fail probability controlled by $\frac{1}{\text{poly}(n)}$. Next, the sequence either reaches $B(X,\delta)$ in $\mathcal{O}(\frac{1}{\alpha\delta^2}\log \frac{1}{\delta})$ iterations, or enters another spurious region $\mathcal{B}(\mathcal{S}_{Z^*},\delta)$ at the time $K_1=\mathcal{O}(\log n)$. In the former case, it further takes $\mathcal{O}(\log \frac{1}{\epsilon})$ iterations to generate an $\epsilon$-accurate solution; in the latter case, by Theorem \ref{thm: stage3}, it takes $\mathcal{O}(\log n)$ iterations to escape $\mathcal{B}(\mathcal{S}_{Z^*},\delta)$, with the fail probability controlled by $\frac{1}{\text{poly}(n)}$. Combining the above results, with high probability exceeding $1-\frac{1}{\text{poly}(n)}$, it takes $\mathcal{O}(\log n+\log\frac{1}{\epsilon})$ iterations for the randomly initialized Riemannian gradient descent (the PGD) to generate an $\epsilon$-accurate solution, i.e. $\|Z-X\|_F<\epsilon\|X\|_F$.
\end{proof}

The proofs of Theorems \ref{thm: stage1}-\ref{thm: stage3} will follow the roadmap laid out in Section \ref{sec: proofskech}. In particular, from Section \ref{sec: DLconverge} to \ref{sec: techlemma}, each subsection will introduce a group of technical results corresponding to a main idea in Section \ref{sec: proofskech}. Specifically, they are the fundamental convergence tool by the \L ojasiewicz inequality, the geometry of spurious regions on $\M_r$, and the dynamics of the trajectory behavior. We then use those technical results to prove Theorems \ref{thm: stage1}-\ref{thm: stage3} in Section \ref{sec: proof3stage}.

\subsection{Fundamental convergence guarantee of the PGD}\label{sec: DLconverge}
The \emph{{\L}ojasiewicz inequality} is a powerful tool for analyzing the convergence rate of gradient-based methods, which is named after S. {\L}ojasiewicz \cite{law1965ensembles,lojasiewicz1963propriete}. Previous works have used the {\L}ojasiewicz inequality to prove the convergence rate in many Euclidean optimization problems as well as Riemannian optimization problems, e.g. \cite{convergence2005absil,bolte2010characterizations,attouch2009convergence,attouch2013convergence,attouch2010proximal,Reinhold,xu2013block}, just to name a few. 

The following theorem serves as a primary tool to determine the convergence rate of the PGD when minimizing a differentiable function $f(\cdot):\M\rightarrow \R$. We assume that $\{Z_k\}$ generated by the PGD is bounded for the rest of the paper.

\begin{theorem}\label{thm: LD2converge}
Let $\M$ be a Riemannian manifold, $f(\cdot):\M\rightarrow \R$ be a differentiable loss function to be minimized, $\{Z_k\}$ ($k=1,2,\ldots$) be a sequence generated by the PGD iterations \eqref{eq: algorithm}. Assume that the following conditions hold:
\begin{enumerate}[label=\arabic*)]
    \item 
    \emph{(Descent Inequality)} There exists $C_d>0$ such that
    \begin{equation}
    \label{eq: D}\tag{D}
        f_k-f_{k+1}\geq C_d \|P_{T_{Z_k}}(\grad f(Z_k))\|\|Z_{k+1}-Z_k\|;
    \end{equation}
    
    \item  
    \emph{({\L}ojasiewicz Gradient Inequality)} There exists $0<C_l<+\infty$ such that 
    \begin{equation}
    \label{eq: L}\tag{L}
        \left|f(Z_k)-f(Z^*)\right|^{1-\omega}\leq C_l\|P_{T_{Z_k}}(\grad f(Z_k))\|,
    \end{equation}
    with $0<\omega\leq \frac{1}{2}$. Here, $Z^*$ is the accumulating point for $\{Z_k\}$.
\end{enumerate}
Then, if the learning rate $\alpha$ satisfies $0<\alpha<\frac{2C_l^2}{C_d}$, the sequence $\{Z_k\}$ converges to their accumulating point $Z^*$ with the following convergence rate:
$$
\|Z_k-Z^*\|\leq
\begin{cases}
e^{-ck} &\mbox{, if $\omega=\frac{1}{2}$};\\
k^{-\frac{\omega}{1-2\omega}} &\mbox{, if $0<\omega<\frac{1}{2}$}.
\end{cases}
$$
When $\omega = \frac{1}{2}$, linear convergence rate can be guaranteed with $c=-\log(1-\frac{\alpha C_d}{2C_l^2})>0$. Here $\|\cdot\|$ is an arbitrary norm under which there is a first order retraction on the manifold.
\end{theorem}
\begin{remark}
A few remarks are in order.
\begin{enumerate}[label=\arabic*)]
    \item This theorem explores the convergence rate of the PGD to an accumulating point. This convergence rate depends on the property of the function $f$ (reflected in the exponent $\omega$), and the constants $C_d>0$, $C_l>0$ and the learning rate $\alpha>0$. 
    \item With Theorem \ref{thm: LD2converge}, the task of checking the convergence rate is reduced to checking conditions (\ref{eq: D}) and (\ref{eq: L}) and determining the respective constants.
    \item This theorem only requires $Z^*$ to be an accumulating point, but $Z^*$ is not necessarily a local minimum. It can also be other types of critical points such as saddle points.
    \item From this result, we can see that the convergence rate is faster with a larger stepsize $\alpha$, or a larger Riemannian gradient $\|P_{T_{Z_k}}(\grad f(Z_k))\|_F$ (which makes $C_l$ smaller).
    \item An extension of this theorem has been proposed in \cite{Sobolev}, which allows for mixed norms in (\ref{eq: D}) and (\ref{eq: L}). 
\end{enumerate}
\end{remark}

\begin{lemma}\label{lem: F12singlestep}
Let $f(Z)=F_1(Z)$ or $F_2(Z)$. Assume that $Z_k$ satisfies condition (D) and condition (L) (defined in Theorem \ref{thm: LD2converge}) \emph{only} for $k=1,\ldots, K$, with $K<+\infty$. Then, we have 
\begin{align*}
    f_{k+1}<(1-\rho)f_k, \quad k = 1,\ldots, K,
\end{align*}
where $f_k:=f(Z_k)$. Further, if $\alpha>0$ is small enough, we have $\|Z_k-X\|_F\lesssim e^{-ck} $, where $\rho=\Omega\left(\alpha\frac{C_d}{C_l^2}\right)\in(0,1)$ and $c=-\frac{1}{2}\log(1-\rho)>0$.
\end{lemma}

The two conditions stated in Theorem \ref{thm: LD2converge} are satisfied in the majority of the manifold $\M_r$ for the loss functions that we consider, including the neighborhood of the ground truth $X$. Thus fast convergence rate in the majority of the manifold is easy to derive. In the rest of the manifold, though, Condition (\ref{eq: L}) could fail and the convergence rate could deteriorate. As we will see in the next section, these regions are the \emph{spurious regions} near the \emph{spurious critical points} on the manifold. Special analysis is needed to study how the PGD escapes these spurious regions. 

We also mention that another version of Theorem \ref{thm: LD2converge} is given in Section \ref{sec: prooftech2}, specially tailored for the objective functions with weak isometry, which will be used in our convergence study of the PGD for phase retrieval.

\subsection{Geometry of the spurious regions on the low-rank matrix manifold}\label{sec: geometry_sr}

In this subsection, we look into the critical points of $F_1(Z) = \frac{1}{2}\|Z-X\|_F^2$, which consists of the ground truth $X$, and the set of spurious critical points denoted as $\mathcal{S}_{Z^*}$. We study the spurious regions $ \mathcal{B}(\mathcal{S}_{Z^*},\delta)$ near those spurious critical points, where Condition (\ref{eq: L}) of Theorem \ref{thm: LD2converge} is violated and special treatment is needed.

\subsubsection{Spurious fixed points on \texorpdfstring{$\overline{\M_r}$}{Mr}}
\begin{definition}[Converging set]
We define the converging set of a fixed point $Z\in\M_r$ as follows:
\begin{align*}
    \mathcal{C}_{Z^*}=\{Z: Z\in \M_r\text{, and for }\{Z_k\}_{k=0}^{\infty}\text{ generated by the PGD initialized at }Z\text{, } \lim_{k\rightarrow \infty}Z_k=Z^*\}.
\end{align*}
\end{definition}

\begin{definition}[Stable/Spurious fixed points]
    For a manifold $\M$ with a Lebesgue measure, we define \emph{spurious fixed points} as the fixed points of an iterative algorithm whose converging sets have zero measure, and \emph{stable} fixed points as those whose converging sets have positive measure.
\end{definition}

\begin{lemma}\label{lem: spurious_fp}
Consider the PGD applied to $F_1(Z)=\frac{1}{2}\|Z-X\|_F^2$. Assume $X = U D V^{\dag}$ is a singular value decomposition of  $X$, where $D \in \R^{r\times r}$ is a non-singular diagonal matrix (the diagonals are not necessarily in descending order), and $U\in \F^{n_1 \times r}$, $V\in\F^{n_2\times r}$. Then, 
\begin{enumerate}[label=\arabic*)]
\item  There are two types of fixed points: one is the ground truth $Z = X$, and the other consists of the set 
\begin{align*}
    \mathcal{S}_{Z^*} := &\big\{Z^*: \, Z^* = U_1D_1 V_1^{\dag}, \text{ where } U = \left(U_1, U_2\right), \, D = \text{diag} \{D_1, D_2\} \text{ and }  V = \left(V_1, V_2\right),\\
    &Z^*\neq X\big\}.
\end{align*}
\item Specifically, if $X$ has distinct singular values\footnote{Which means that the eigenvalues of $X$ all have algebraic multiplicity equal to 1.}, then $\mathcal{S}_{Z^*}$ has cardinality $|\mathcal{S}_{Z^*}|=2^r-1$. Assume that $X=\sum_{i=1}^r d_i u_i v_i^{\dag}$, then $\mathcal{S}_{Z^*}=\{Z^*=\sum_{i=1}^r d_i\eta_i u_i v_i^{\dag}, \text{with }\eta\in\{0,1\}^r \text{ and } \eta\neq (1,\,1,\,\ldots,1)^{\dag}\}$.
\end{enumerate}
\end{lemma}

The set $\mathcal{S}_{Z^*}$ consists of $2^r-1$ points (including $Z^*=0$) if the singular values of $X$ are distinct, or contains some submanifolds of $\M_r$ if at least one singular value has multiplicity more than one. In Theorem \ref{thm: rankrconverge}, one can show that with high probability, $\mathcal{S}_{Z^*}$ is the set of spurious fixed points of the PGD, while $Z=X$ is a stable fixed point. We call them ``spurious fixed points'' because  they are not stable and when the tangent cone is taken into consideration they are not true fixed points. Moreover, with high probability the sequence generated by the randomly initialized PGD does not converge to any one of these spurious fixed points. Below is an example.

\begin{example}
\label{ex: spurious}
    Assume that $n=3$, $r=2$. Let 
    \begin{align*}\small
        X=\begin{pmatrix}1&0&0\\0&1&0\\0&0&0\end{pmatrix}, \qquad Z_0=\begin{pmatrix}1&0&0\\0&0&0\\0&0&1\end{pmatrix}.
    \end{align*}
    Then the $\{Z_k\}_{k=0}^\infty$ generated by the PGD and their limit point are given by
    \begin{align*}\small
        Z_k=\begin{pmatrix}1&0&0\\0&0&0\\0&0&(1-\alpha)^k\end{pmatrix}, \qquad Z^*:=\lim_{k\to\infty}Z_k=\begin{pmatrix}1&0&0\\0&0&0\\0&0&0\end{pmatrix}.
    \end{align*}
    We see that $Z^*$ is a spurious critical point. Note that even though each $Z_k$ is in $\M_2$, their limit is in $\M_1$. \\
    Figure \ref{fig: Zstararea} is a visualization of the gradient $\|P_{T_{Z}}(Z-X)\|_F$ in the neighborhood of a spurious $Z^*$. We can see that the gradient is essentially singular near $Z^*$. There is only one direction in which the sequence converges to $Z^*$. Along other directions, the Riemannian gradient remains large and the PGD will likely slip away. We point out that such property of these spurious fixed points is very similar to that of \emph{strict saddle points} in many nonconvex optimization problems, although they are not exactly the same.
    \begin{figure}[ht]
        \centering
        \includegraphics[width=0.4\textwidth]{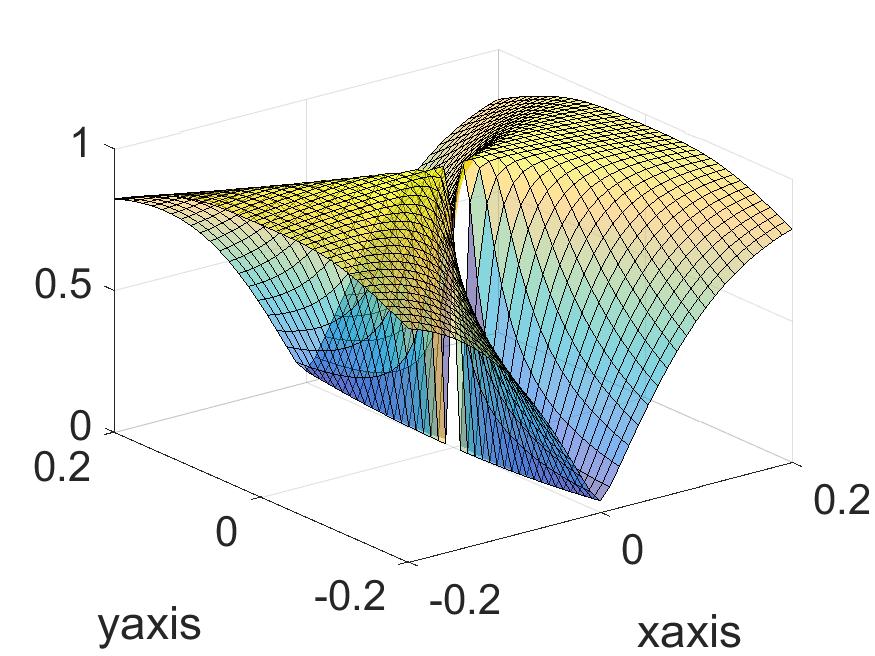}
        \caption{Magnitude of the gradient in the neighborhood of a spurious fixed point}\label{fig: Zstararea}
    \end{figure}
\end{example}

Example \ref{ex: spurious} demonstrates the importance of considering $\overline{\M_r}$ instead of $\M_r$ itself, and gives us a healthy warning that we cannot assume the PGD always stays in the interior of $\M_r$ and converges to a minimum. 

However, numerical evidence shows that for the majority of cases, the PGD almost surely escapes the spurious critical points $\mathcal{S}_{Z^*}$ and converges to the global minimum. In fact, we will prove that with high probability the set of initial points that converge to $\mathcal{S}_{Z^*}$ is a measure-zero lower-dimensional submanifold on $\M_r$.

\subsubsection{Spurious regions in the neighborhood of spurious fixed points} 
As mentioned in Section \ref{sec: DLconverge}, there are some regions on $\M_r$ that violate the conditions for fast convergence guarantee in Theorem \ref{thm: LD2converge}. In this subsection, we take a detailed look at those regions. 

To ensure that Condition (\ref{eq: L}) holds with $\omega=\frac{1}{2}$, what we essentially need is
$$\|P_{T_z}(Z-X)\|_F \ge C_L, \qquad  C_L>0.$$ 
However, for some special $Z$ which occupies a small part of the whole domain, the Riemannian gradient $P_{T_z}(Z-X)$ becomes so small that this lower bound is violated. We use 
\emph{spurious regions} to refer to the regions where $Z$ violates this lower bound.

The following results show that the spurious regions are in the neighborhood of the spurious fixed points. 
\begin{lemma}[Spurious regions]
\label{lem: balls}
    Assume that $X = U D V^{\dag}$ is the singular value decomposition of $X$. Let $0<\delta < d_r/2$. Then the spurious regions can be charaterized as follows
    \begin{align*}
        \mathcal{B}(\mathcal{S}_{Z^*},\delta) & := \{Z: \|P_{T_z}(Z-X)\|_F \le \delta\} \cap \M_r\cap\{Z:\|Z-X\|_F>\frac{d_r}{2}\}\\
        & = \left\{Z: \, Z = 
		\left(
			B , \widetilde{B}
		\right)
		\begin{pmatrix}
			D_1 + E_1 & 0  \\
			0 & E_2 
		\end{pmatrix}
		\begin{pmatrix}
			C^{\dag} \\
			\widetilde{C}^{\dag}
		\end{pmatrix}, \right. \\
		& \|E_1\|,\, \|E_2\|,\, \|P_{B} - P_{U_1}\|,  \|P_{\widetilde{B}} - P_{\widetilde{U}}\|, \|P_{C} - P_{V_1}\|,  \|P_{\widetilde{C}} - P_{ \widetilde{V}}\| \leq \mathcal{O}(\delta)\bigg\}\\
		&=\cup_{Z^*\in\mathcal{S}_{Z^*}} \mathcal{B}(Z^*,\delta)
    \end{align*}
    where $U = (U_1, U_2)$ and $V = (V_1, V_2)$ are the $(s, r-s)$ dimensional splitting for some $0<s<r$; $\widetilde{U}$, $\widetilde{V}$: $\tilde{U}\subset col(U)^{\perp}$, $\widetilde{V}\subset col(V)^{\perp}$,  $\widetilde{U}^{\dag}\widetilde{U}=\widetilde{V}^{\dag}\widetilde{V}=I_{r-s}$, and $D=\text{diag}(D_1,D_2)$ where $D_1$ and $D_2$ are diagonal matrices.
\end{lemma}

\begin{remark}\label{def: spuriousfp}
    The intuition behind Lemma \ref{lem: balls} is that an $s$-dimensional principal part of $Z$ is ``almost aligned'' with an $s$-dimensional principal part of $X$, and their singular values are close to each other; while the other ($r-s$)-dimensional part of $Z$ is ``almost perpendicular'' to the other ($r-s$)-dimensional part of $X$, and the singular values of that part of $Z$ is very small. 
\end{remark}

\begin{lemma}\label{lem: ness}
For the case when the matrices are symmetric positive semi-definite (SPSD), there exist eigenvalue decompositions $X=UDU^{\dag}$, $Z=V_z\Sigma_z V_z^{\dag}$ (diagonals of $\Sigma_z$ are not required to be in descending order) and $U=V_zR+\widetilde{V}_z S$, with $\widetilde{V}_z\in Col(V_z)^{\perp}$. If $Z\in \mathcal{B}(\mathcal{S}_{Z^*},\delta)$, where $\text{rank}(Z^*) = s$, then we have:
\begin{align*}
    R&=\begin{pmatrix}
    I_s+E_1 & E_2\\
    E_3 & E_4
    \end{pmatrix},
\end{align*}
with $\|E_1\|^{\frac{1}{2}}$, $\|E_2\|$, $\|E_3\|$, and $\|E_4\|\leq \mathcal{O}(\delta)$, rank$(Z^*)=s$, $0<s<1$, $S\in\F^{r\times r}$. 
\end{lemma}
\begin{remark}
    In the case where the eigenvalues of $X$ are not distinct, simply let $U_1$ be the basis of the best subspace that $V_z$ can capture.
\end{remark}

\begin{definition}[Intrinsic/Spurious branch]\label{def: intr_sp_branch}
We say that a point $Z_0\in\M$ belongs to an \emph{intrinsic branch} if the limit point $\lim_{k\rightarrow\infty}Z_k$ is a stable fixed point, where $\{Z_k\}$ is generated by the PGD with initialization at $Z_0$. We say that $Z_0$ belongs to a \emph{spurious branch} if the limit point of $\{Z_k\}$ is a spurious fixed point.
\end{definition}

\begin{remark}
    The intrinsic branch corresponds to the set of good initial points in Theorem \ref{thm: rankrconverge} which can be sampled with high probability no less than $1-\frac{1}{\poly(n)}$. The spurious branch, on the other hand, has probability measure no more than $\frac{1}{\poly(n)}$. We will improve the bound on some of the spurious branches to zero measure in an upcoming work.
\end{remark}

\subsubsection{Convergence rate outside the spurious regions}
We now show that as long as the spurious regions are excluded, we can establish the linear convergence rate of the PGD, using Conditions (\ref{eq: D}) and (\ref{eq: L}) in Theorem \ref{thm: LD2converge}.  

\begin{figure}[ht]
        \centering
        \includegraphics[width=0.4\textwidth]{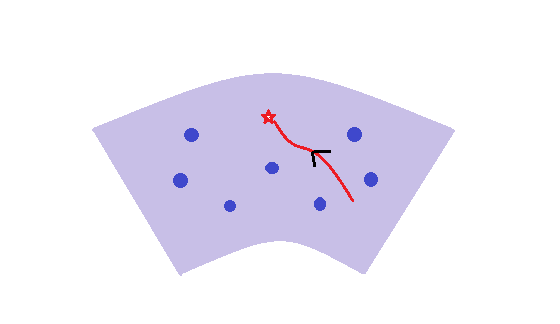}
        \caption{Convergence guarantee on the majority of $\M_r$}\label{fig: majarea}
\end{figure}

\begin{lemma}\label{lem: LS}
Let $F_1(Z)=\frac{1}{2}\|Z-X\|_F^2$, $X, Z\in\M_r$. If the $Z_k$'s generated by the PGD remain in a bounded subset of $\M_r$, and stay in the set $\{Z: \|P_{T_z}(Z-X)\|_F\geq C_L\}\cup \{Z:\|Z-X\|_F\leq\frac{d_r}{2}\}$, if $\alpha>0$ is a properly small constant, we have: 
\begin{enumerate}[label=\arabic*)]
        \item $\|P_{T_z}(\grad f(Z_k))\|_F\geq C_1\|Z_k-X\|_F$, for all $k$, with $C_1\geq \Omega(C_L)>0$. That is, condition (\ref{eq: L}) holds with $\omega=\frac{1}{2}$;
        \item There exists some absolute constant $C_2>0$ such that
            \begin{align*}
                f_k-f_{k+1}\geq  C_2 \|P_{T_{Z_k}}(\grad f(Z_k))\|_F \| Z_{k+1}-Z_k\|_F.
            \end{align*}
    \end{enumerate}
    Thus, by Theorem \ref{thm: LD2converge}, there exists $\alpha>0$ such that the sequence $\{Z_k\}$ generated by the PGD (\ref{eq: algorithm}) converges to $X$ in a linear convergence rate:
    $$
        \|Z_k-X\|_F\leq e^{-ck}.
    $$
Here, $c=-\log (1-\Omega(\alpha C_L^2))>0$.
\end{lemma}
Lemma \ref{lem: LS} ensures linear convergence to $X$ in the majority of the manifold outside of the spurious regions. 
The rest of the analysis thus evolves around how the trajectory of the PGD escapes the spurious regions.

\subsection{Technical Lemmas}\label{sec: techlemma}
In Section \ref{sec: KIS}, we have outlined how the trajectory of the PGD is divided into three stages, corresponding to Theorem \ref{thm: stage1}, Theorem \ref{thm: stage2} and Theorem \ref{thm: stage3}. In this subsection, we introduce a few technical lemmas needed for the proofs of Theorems \ref{thm: stage1}-\ref{thm: stage3}.

We first define a general random distribution on the low-rank manifold.

\begin{definition}[General random distribution]\label{def: GRD}
$Z$ is said to be drawn from a \emph{general random initialization},  if $Z=V_1\Sigma V_2^{\dag}$ where $V_1$ and $V_2$ are drawn from a uniform distribution on the Stiefel manifold $\mathbb{V}_r(\R^n)$, and the entries of $\Sigma$ are drawn independently from a uniform distribution over $[C_1,C_2]$ with $C_2>C_1\geq 0$.
\end{definition}
\begin{remark}
\begin{enumerate}[label=\arabic*)]
    \item The simplest example of a general random distribution is the following rank-1 Gaussian sampling. One can construct $Z_0=c u_0u_0^{\dag}\in\mathbb{F}^{n\times n}$, where $c>0$ is a constant and $u_0$ is drawn from $\frac{1}{\sqrt{n}}\mathcal{N}(0,1)^n$ for $\F=\R$, or $\frac{1}{\sqrt{2n}}\mathcal{N}(0,1)^n+i\frac{1}{\sqrt{2n}}\mathcal{N}(0,1)^n$ for $\F=\C$. It is equivalent to $Z_0=\rho v_0v_0^{\dag}$ with $v_0$ drawn from $\mathbb{V}_1(\R^n)$ and $\rho$ drawn from $\Omega(\frac{1}{n})\chi^2(n)$ or $\Omega(\frac{1}{2n})\chi^2(2n)$.\footnote{Defined as $\chi^2(s):=\sum_{j=1}^s n_j^2$ where $n_j$, $j=1,2,...,s$ are i.i.d standard normal variables.}
    \item In practical computation, a uniform distribution on the Stiefel manifold can be easily constructed as follows. For $\F=\R$, let $G$ be drawn from $\frac{1}{\sqrt{n}}\mathcal{N}(0,1)^{n\times r}$, and construct $V=G(G^{\dag}G)^{-\frac{1}{2}}$. For $\F=\C$, change the law of $G$ to $\frac{1}{\sqrt{2n}}\left(\mathcal{N}(0,1) + i\cdot\mathcal{N}(0,1)\right)^{n\times r}$.
   \item The density functions of the sampling laws of $V$ and $\Sigma$ can be extended from constants to more general ones. For example, it suffices to require that $\rho(V) \in [c,C]$ for any $V\in \mathbb{V}_r (\R^n)$, where $C>c>0$. This allows more flexibility in the initialization.
\end{enumerate}
\end{remark}
For any given $s\in[r]$, we have that the marginal distribution of $V_s:=V(:,s)$ is an uniform distribution on $\mathcal{S}^{n-1}$. Therefore, for any given $u:\|u\|_2=1$, we have $\|V_s^{\dag}u\|\gtrsim \frac{1}{\sqrt{n}}$ with high probability, if we ignore the log-factor. Specifically, we have the following lemma.

\begin{lemma}\label{lem: uniformST}
Assume $W=(W_1,W_2,...,W_r)\sim \text{Uniform }(V_r(\R^n))$. For any $u_0\in\F^n$ such that $\|u_0\|_2=1$, we have:
\begin{enumerate}[label=\arabic*)]
    \item $\mathbb{E}(\|u_0^{\dag}W\|_2^2)=\frac{r}{n}$;
    \item $\text{Prob}(|u_0^{\dag}W_i|^2\geq\Omega(\frac{1}{n\log n}))\geq 1-e^{-\Omega(n)}- \mathcal{O}(\frac{1}{\log n})$, $i\in[r]$;
   \item $\text{Prob}(|u_0^{\dag}W_i|^2\geq \Omega (\frac{\log n}{n}))\leq \frac{1}{\text{poly}(n)}+e^{-\Omega(n)}$, $i\in[r]$;
    \item $\text{Prob}(|u_0^{\dag}W_i|^2\geq\Omega(\frac{1}{n^{p+1}}))\geq 1-e^{-\Omega(n)}- \mathcal{O}(\frac{1}{n^p})$, $i\in[r]$.
\end{enumerate}
\end{lemma}

Lemma \ref{lem: uniformST} shows that the general random distribution with high probability captures weak information of order $\Omega(\frac{1}{n})$ of a given column space if we ignore some log factors.

The following is an important technical result that describes the dynamics of the singular values and column spaces of the iterative sequence. 

\begin{lemma}\label{lem: Rdynamics}
Consider the gradient flow of $F_1(Z)=\frac{1}{2}\|Z-X\|_F^2$. Assume $Z=V_z\Sigma_zV_z^{\dag}$ (diagonals of $\Sigma_z$ are not required to be in descending order) and $X=UDU^{\dag}$ are the eigenvalue decompositions of the SPSD (symmetric positive semidefinite) matrices $Z$ and $X$ respectively. Denote $U=V_zR+\widetilde{V}_zQ$ with $\widetilde{V}_z\in Col(V_z)^{\perp}$. We have:
$$\frac{d}{dt}\sum_{j=1}^r\|D^{\frac{1}{2}}R^{\dag}e_j\|^2=\sum_{j=1}^r\frac{2}{\Sigma_z(j,j)}\|QDR^{\dag}e_j\|^2.$$ 
Furthermore, denote the spectra of $R^{\dag}R$ and $RR^{\dag}$ by $\Sigma_{RR}=\text{diag}\{r_1,r_2,...,r_r\}$, where $r_1\ge\ldots\ge r_r$, then we have:
\begin{align}\label{eq:sumr}
    \frac{d}{dt}(\sum_{j=1}^r r_j)\gtrsim \sum_{j=1}^r(1-r_j)r_j,
\end{align}
and
\begin{align*}
    \frac{d}{dt}(\Sigma_z(j,j))&=(RDR^{\dag})(j,j)-\Sigma_z(j,j).
\end{align*}
Further, if $\sigma_j=\Omega(1)$ for all $j\in[r]$, then we have $\frac{d}{dt}\sum_{j=1}^r r_j =\Omega\left(\sum_{j=1}^r (1-r_j)r_j\right)$.
\end{lemma}
\begin{remark}
For all $j\in[r]$, we have $0\leq r_j\leq 1$, where the $r_j$'s describe the angle between the column spaces of $Z$ and $X$. From (\ref{eq:sumr}), we can conclude that $\sum_{j=1}^r r_j$ is non-decreasing. The increase of $\sum_{j=1}^r r_j$ only slows down if for all $j\in[r]$, $r_j$ is close to either 0 or 1. Note that $\sum_{j=1}^r r_j$ is an indicator of whether the sequence is close to a spurious critical point. For example, when $Z \in \mathcal{B}(0,\delta)$ and $\sum_{j=1}^r r_j = \Omega(\delta)>0$, we have $\frac{d}{dt}(\sum_{j=1}^r r_j) \ge C \cdot \sum_{j=1}^r r_j$ for some constant $C$, i.e. the sum increases exponentially fast. The sequence eventually leaves $\mathcal{B}(0,\delta)$ and never comes back. The other spurious critical points $Z^* \neq 0$ can be treated similarly, see the proof of Theorem \ref{thm: stage3} in Section \ref{sec: proof3stage}.
\end{remark}

To better illustrate the idea for the proof of Theorem \ref{thm: stage3}, i.e. the dynamics of the third stage, here is a toy example in the Euclidean space that demonstrates the pull-back of the projection along certain coordinates. \footnote{In an earlier version of this paper, we used a slightly different idea based on the pull-back of \emph{volumes}. We opt for the pull-back of \emph{projections} in this version for the simplicity of the proof.}    

\begin{example}\label{eg: euclieg}
Consider using the gradient descent to minimize
\begin{align*}
f(z)&=z^{\dag}
\begin{pmatrix}
I_{n-s}&0\\0&-I_s    
\end{pmatrix} z.
\end{align*}
We have
\begin{align*}
    z_{k+1}=\begin{pmatrix}
(1-\alpha)I_{n-s}&0\\0&(1+\alpha)I_s
\end{pmatrix} z_k.
\end{align*}
Obviously, $z^*=0$ is a saddle point. Define $B=B(0,\delta)$ as the $\delta$-ball centered at 0, for some small $\delta>0$ and any $j$ satisfying $n-s+1\leq j\leq n$, we have $e_j^{\dag}\Phi_{\alpha}^{-K}(e_j)\leq \left(1-\Omega(\alpha)\right)^K$. For any $y\in B$, assume $y=y_je_j+\text{Col}(e_j)^{\perp}$. We consider $e_j^{\dag}\Phi_{\alpha}^{-K}(y_je_j+\text{Col}(e_j)^{\perp})=y_j\left(1-\Omega(\alpha)\right)^K$. If we take $K=\Omega(\log n)$, we have $e_j^{\dag}\Phi_{\alpha}^{-K}(y)\lesssim\frac{1}{\poly (n)}$. If $z_0$ is sampled from a random initialization in a bounded region, by Lemma \ref{lem: uniformST}, we have $\mu(\Phi_{\alpha}^{-K}(B))\lesssim \frac{1}{\poly(n)}$.
\end{example}

Finally, the following lemma gives the local {\L}ojasiewicz inequality in the neighborhood of the ground truth $X$ as well as the local convergence rate, which is used in the proof of Theorem \ref{thm: rankrconverge}.

\begin{lemma}[Local convergence]\label{lem: P_Tnorm}
For $F_1(Z)=\frac{1}{2}\|Z-X\|_F^2$, let $Z = U_z \Sigma V_z^\dag$ and $X = U D V^\dag$. Denote $\sigma_j$ and $d_j$ as the $j$-th largest singular value of $Z$ and $X$ respectively.
\begin{enumerate}[label=\arabic*)]
    \item  We have
        \begin{align*}
            \frac{\sigma_r^2}{\sigma_r^2+\|X\|_F^2}\|Z-X\|_F^2\leq \|P_{T_z}\left(Z-X\right)\|_F^2\leq \|Z-X\|_F^2;
        \end{align*}
    \item In the local region around the ground truth $\{Z: \text{rank}(Z)=r,\, \|Z-X\|_F<\frac{d_r}{2}\}$, we have the lower bound of $\sigma_r$ as  $\sigma_r>\frac{d_r}{2}>0$ and $\frac{\|P_{T_z}(Z-X)\|_F}{\|Z-X\|_F}\gtrsim d_r$. Further, if stepsize $\alpha>0$ is properly small, we have $\|\phi_{\alpha}(Z)-X\|_F \leq e^{-c}\|Z-X\|_F$ with $c=-\log\left(1-\Omega(\alpha d_r^2)\right)>0$.
\end{enumerate}
\end{lemma}

\subsection{Proofs of Theorem \ref{thm: stage1}-Theorem \ref{thm: stage3}}
\label{sec: proof3stage}

In this subsection, we give the detailed proofs of  Theorems \ref{thm: stage1}-\ref{thm: stage3} using the technical results from the previous three subsections. Here, we assume $R=U^{\dag}V_z$ satisfies the Assumption 3 and note there are some properties derived from Lemma \ref{lem: Rdynamics}.

\begin{proof}[\textbf{Proof of Theorem \ref{thm: stage1}}]
Our goal is to show that with only a small probability, the sequence starting from $Z_0$ will reach $\cup_{Z^*\in\mathcal{S}_{Z^*}\setminus \{0\}}\mathcal{B}(Z^*,\delta)$ within $K_0=\mathcal{O}(1)$ number of iterations. By Lemma \ref{lem: uniformST}, with the fail probability controlled by $\frac{1}{\poly(n)}$, we have $\|R\|_F\lesssim \frac{\log n}{n}$, where $R =  V_z^\dag U$. This implies $\sum_{j=1}^r r_j \lesssim \frac{\log n}{n}$. Since $Z_0$ is drawn from the general random distribution, we have $\sigma_j=\Omega(1)$, for all $j\in[r]$. By Lemma \ref{lem: Rdynamics}, we have $\frac{d}{dt}\sigma_j=(RDR^{\dag})(j,j)-\sigma_j\geq -\sigma_j$. With Assumption 1, we have for all $k\leq K_0=\Omega(1)$, $\sigma_j=\Omega(1)$. Further, $\frac{d}{dt}\sum_{j=1}^r r_j=\Omega(\sum_{j=1}^rr_j(1-r_j))$. Therefore, for $k\leq K_0=\mathcal{O}(1)$, with high probability exceeding $1-\frac{1}{\poly(n)}$, we have $\sum_{j=1}^r r_j\lesssim \frac{\log n}{n}\leq \frac{1}{2}$. By Lemma \ref{lem: ness}, this implies that the sequence will not reach any $\mathcal{B}(Z^*,\delta)$ with $Z^*\in\mathcal{S}_{Z^*}\setminus \{0\}$ in $\mathcal{O}(1)$ number of iterations, if we require $\delta>0$ small enough.
\end{proof}

\begin{proof}[\textbf{Proof of Theorem \ref{thm: stage2}}]
Following the proof of Theorem \ref{thm: stage1}, at $k=K_0$, with fail probability controlled by $\frac{1}{\poly(n)}$, we have $\frac{1}{\poly(n)}\lesssim \|R\|_F\lesssim \frac{\log n}{n}$. Therefore,
in the region $\mathcal{B}(0,\delta)$, by Lemma \ref{lem: Rdynamics}, we have $\frac{d}{dt}\sum_{j=1}^r r_j\gtrsim \sum_{j=1}^r r_j$. With Assumption 1, we have $\sum_{j=1}^r r_j\gtrsim \delta$ at step $K_1=K_0+C_1\log n$ with some $C_1>0$. By Lemma \ref{lem: ness}, this implies that the sequence has escaped from $\mathcal{B}(0,\delta)$. Since $\sum_{j=1}^r r_j$ is non-decreasing, $\sum_{j=1}^r r_j \gtrsim \delta$ for all $k\geq K_1$, so the sequence will not come back to $\mathcal{B}(0,\delta)$. 
\end{proof}

\begin{proof}[\textbf{Proof of Theorem \ref{thm: stage3}}]
Let $S^c \subset [r]$ be the index set such that $Z^*=X-\sum_{j\in S^c}d_jU_jU_j^{\dag}$. For $j \in S^c$, without loss of generality, we have $R(j,j)\geq 0$ for $j\in[r]$. In the sequel, let $R(j,j)^+$ always denote the value of $R(j,j)$ one step after the current step.  By (\ref{eq: Rdy}) in the proof of Lemma \ref{lem: Rdynamics}, we have
\begin{align*}
    \Delta R(j,j)&:=R(j,j)^{+}-R(j,j)\\
    &=\alpha\cdot  e_j^{\dag}M^{\dag}Re_j+\alpha\cdot \frac{1}{\sigma_j}e_j^{\dag}RDQ^{\dag}Qe_j+o(\alpha)\\
    &=\alpha \cdot \sum_{l\neq j}M(l,j)R(l,j)+\alpha\cdot \frac{d_j}{\sigma_j}R(j,j)Q^{\dag}Q(j,j)+\alpha\cdot \sum_{l\neq j} \frac{d_l}{\sigma_j}R(j,l)Q^{\dag}Q(l,j)
    +o(\alpha).
\end{align*}
By Assumption 2, there exists a constant $c_g$ such that $\min_{i\neq l}|\sigma_i - \sigma_l| \ge c_g \sigma_1$. Using equation (\ref{eq: M}) in the proof of Lemma \ref{lem: Rdynamics} and Assumption 3, we have
\begin{align*}
    M(i,j)&=\frac{(RDR^{\dag})(i,j)}{\sigma_j-\sigma_i}\\
    &\leq \frac{d_1\|R^{\dag}e_i\|\|R^{\dag}e_j\|}{c_g\sigma_1}\\
    &\leq \frac{d_1\left(\|R^{\dag}e_i\|^2+\|R^{\dag}e_j\|^2\right)}{2c_g\sigma_1}\\
    & \le  \frac{d_1 \|R^{\dag}e_i\|^2}{2c_g\sigma_i} +  \frac{d_1 \|R^{\dag}e_j\|^2}{2c_g\sigma_j}\\
    &\leq \frac{d_1(RDR^{\dag})(i,i)}{2d_rc_g\sigma_i}+ \frac{d_1(RDR^{\dag})(j,j)}{2d_rc_g\sigma_j}\\
    &\leq \frac{d_1C_u}{d_rc_g}.
\end{align*}
Thus there exists a constant $C_M = \mathcal{O}(1) >0$ such that $\|M\|_F\leq C_M$. Using the relation $R^{\dag}R+Q^{\dag}Q=I_r$ and Assumption 1, we have that
\begin{align*}
    \Delta R(j,j)&\geq \alpha\cdot \frac{d_j}{\sigma_j}R(j,j)(1-\|Re_j\|^2)
    -\alpha\cdot c_1 R(j,j)-\alpha\cdot \frac{c_2}{\sigma_j} \cdot R(j,j)^2\\
    &=\alpha\cdot \left( \frac{d_j(1-\|Re_j\|^2)-c_2R(j,j)}{\sigma_j}-c_1     \right)\cdot R(j,j)\\
    &\geq \alpha\cdot \left(\frac{d_j(1-rL^2R(j,j)^2)-c_2R(j,j)}{\sigma_j}-c_1\right)\cdot R(j,j),
\end{align*}
where $c_1=\sqrt{r}C_ML$ and $c_2=rL^2\cdot \sum_{l\neq j}d_l$ are positive constants. On the other hand, we have
\begin{align*}
    |\Delta R(j,j)|\leq \alpha\cdot R(j,j)\cdot \left(\frac{d_j+c_2}{\sigma_j}+c_1\right).
\end{align*}
We now consider two cases: 
\begin{itemize}
    \item \textbf{Case 1:} $\sigma_j > \sigma_j^*=\min\left\{\frac{d_j}{2rL^2C_u},\frac{d_j^3}{9c_2^2C_u},\frac{d_j}{6(1+c_1)} \right\}$.

In this case, we require $$\alpha\leq \frac{2\sigma_j^*}{3\left(d_j+c_2+c_1\sigma_j^*\right)} = \Omega(1).$$
This gives  
$$|\Delta R(j,j)|\leq \frac{2}{3}R(j,j),$$
and 
$$R(j,j)^+\geq \frac{1}{3}R(j,j).$$
    \item \textbf{Case 2:} $\sigma_j \le \sigma_j^*=\min\left\{\frac{d_j}{2rL^2C_u},\frac{d_j^3}{9c_2^2C_u},\frac{d_j}{6(1+c_1)} \right\}$.
 
By Assumption 3, we have $R(j,j)\leq \sqrt{\frac{C_u\sigma_j}{d_j}}$. If we have $\sigma_j\leq \min\left\{\frac{d_j}{2rL^2C_u},\frac{d_j^3}{9c_2^2C_u},\frac{d_j}{6(1+c_1)} \right\}$, then we have
\begin{align*}
    \Delta R(j,j)\geq \alpha R(j,j),
\end{align*}
and 
\begin{align*}
    R(j,j)^+\geq (1+\alpha)R(j,j).
\end{align*}
Note that in this case we do not impose any extra condition on the step size $\alpha$.
\end{itemize}

Now consider $H_{\delta}:=\{u: |u^{\dag}U_j|<\delta\}$, since for any $Z\in \mathcal{B}(Z^*,\delta)$ and $j\in S^c$, by Lemma \ref{lem: ness}, we have $V_{z,j}\in H_{\delta}$. Using the relation between $R(j,j)$ and $R(j,j)^{+}$, for Case 1, have
\begin{align*}
     \max_{u\in H_{\delta}}|U_j^{\dag}\Phi_{\alpha,j}^{-1}(u)|\leq 3\delta.
\end{align*}
Therefore, $\Phi_{\alpha, j}^{-1}(H_{\delta})\subset H_{3\delta}$.
On the other hand, for Case 2, we have
\begin{align*}
     \max_{u\in H_{\delta}}|U_j^{\dag}\Phi_{\alpha,j}^{-1}(u)|\leq \frac{\delta}{1+\alpha}.
\end{align*}
Therefore, we have
$\Phi_{\alpha, j}^{-1}(H_{\delta})\subset H_{\frac{\delta}{1+\alpha}}\subset H_{\delta}$.

We are ready to compute the following quantity, which is an upper bound on the measure of the set of initial points, from which the sequence is trapped by $H_{\delta}$ (i.e., trapped by the corresponding spurious regions) by at least $C_2\log n$ steps:
\begin{align*}
     &\mu\left(\Phi_{\alpha,j}^{-K_0-\mathcal\mathcal{O}(\log n)-K_1-C_2\log n}(H_{\delta})\right) .
\end{align*}
If this measure is small, then with high probability, the number of iterations that the sequence stays in $\mathcal{B}(\mathcal{S}_{Z^*}\setminus\{0\},\delta)$ is no more than $C_2\log n$ iterations, where $C_2=\Omega(1)>0$ is a constant.

Now, during Stage 1 when the iteration $k\leq K_0$, we have $\sigma_r\geq \Omega(1)$, thus $\Phi_{\alpha,j}^{-1}(H_{\delta})\subset H_{3\delta}$. During Stage 2 when $K_0\leq k\leq K_0+\mathcal{O}(\log n)$, we have $\sigma_r$ and $R(r,r)$ are both small enough, if requiring $\delta>0$ small enough. Therefore, we have $\Phi_{\alpha,j}^{-1}(H_{\delta})\subset H_{\frac{\delta}{1+\alpha}}$. 

Afterwards, the sequence either stays inside or outside of spurious regions. In either case, we have $\Phi_{\alpha,j}^{-1}(H_{\delta})\subset H_{3\delta}$ as long as $\alpha>0$ is smaller than an absolute constant. Note that $K_1$ denotes the number of steps spent outside of spurious regions and outside of the local neighborhood $B_\delta(X)$. In fact, $K_1 = \mathcal{O}(\log(1/\delta))$. This is because by Lemma \ref{lem: LS}, $f_k$ is monotone decreasing. Thus it takes no more than $\frac{2}{c}\log(\|Z_0-X\|_F/\delta)$ steps outside of spurious regions before the sequence reaches the local neighborhood $B_\delta(X)$, where $c$ is the constant in Lemma \ref{lem: LS}.

Combining all the above, we have
\begin{align*}
      &\mu\left(\Phi_{\alpha,j}^{-K_0-\mathcal\mathcal{O}(\log n)-K_1-C_2\log n}(H_{\delta})\right) \\
      &\leq \mu\left(\Phi_{\alpha,j}^{-K_0-\mathcal\mathcal{O}(\log n)-K_1}(H_{\frac{\delta}{(1+\alpha)^{C_2\log n}}})\right) \\
      & \leq \mu\left(\Phi_{\alpha,j}^{-K_0-\mathcal\mathcal{O}(\log n)}(H_{\frac{\delta}{(1+\alpha)^{C_2\log n}}\cdot 3^{K_1}})\right)\\
      & \leq  \mu\left(\Phi_{\alpha,j}^{-K_0}(H_{\frac{\delta}{(1+\alpha)^{C_2\log n+\mathcal{O}(\log n)}}\cdot 3^{K_1}})\right)\\
      &\leq  \mu\left(H_{\frac{\delta}{(1+\alpha)^{C_2\log n+\mathcal{O}(\log n)}}\cdot 3^{K_1+K_0}}\right).
\end{align*}
By requiring $C_2>0$ to be a large enough constant, using Lemma \ref{lem: uniformST} (4), we have $$\mu\left(\Phi_{\alpha,j}^{-K_0-\mathcal\mathcal{O}(\log n)-K_1-C_2\log n}(H_{\delta})\right)\leq \frac{1}{\poly(n)}.$$
\end{proof}

\section{Proofs of Theorem \ref{thm: rank1thm} and Theorem \ref{thm: pr}}
\label{sec: prooftech2}
In this section we will prove the Main result 2 (Theorem \ref{thm: rank1thm}) and Main result 3 (Theorem \ref{thm: pr}) mentioned in Section \ref{sec: mainresults}. Theorem \ref{thm: rank1thm} is a special case of Theorem \ref{thm: rankrconverge} restricted to $r=1$, and its proof is much simpler. Theorem \ref{thm: pr} builds upon the previous theorem, but extends the analysis to the case of weak isometry. We provide some insights on the connections between Theorem \ref{thm: rank1thm} and Theorem \ref{thm: pr} in Section \ref{sec: connection23}.  

\subsection{Convergence tool for weak isometry}
We first introduce Theorem \ref{thm: Converge12}, a variant of Theorem \ref{thm: LD2converge}, as a fundamental tool for analyzing the convergence rate for functions with weak isometry as in Theorem \ref{thm: pr}. Using this theorem, we can show that if the measurement sampling operator preserves the distances of points on the manifold to the ground truth $X$ to some extent (indicated by $C_1$ and $C_2$ in the distance-preserving condition below), and the projection operator satisfies a similar property as before, then with this $T(\cdot)$ operator, the loss function $f(Z)=\frac{1}{2}\|T(Z)-T(X)\|_F^2$ still preserves the nice properties of the original least squares loss function $F_1(Z)=\frac{1}{2}\|Z-X\|_F^2$. As a result, the sequences generated by the PGD still converge to the ground truth in a linear rate on the manifold as long as they stay outside of the spurious regions.

\begin{theorem}\label{thm: Converge12}
Assume $T:\mathcal{M}_r\rightarrow \R^m$ is a linear operator, and $f(Z)=\frac{1}{2}\|T(Z)-T(X)\|_2^2$. If the following conditions hold:
    \begin{enumerate}[label=\arabic*)]
    \item \emph{(Distance-preserving condition)} $C_1\|Z_k-X\|_F\leq\|T(Z_k)-T(X)\|_2\leq C_2\|Z_k-X\|_F$ and $C_1\|Z_{k+1}-Z_k\|_F\leq\|T(Z_{k+1})-T(Z_k)\|_2\leq C_2\|Z_{k+1}-Z_k\|_F$, where $C_1$, $C_2>0$ are uniform constants for all $k$; 
    \item \emph{(Critical ratio condition)} $\|Z_k-X\|_F\leq C_3\|P_{T_{Z_k}}(\grad f(Z_k))\|_F$, where $C_3>0$ is a positive constant for all $k$.
    \end{enumerate}
    Then, Conditions (\ref{eq: D}) and (\ref{eq: L}) hold with $\omega=\frac{1}{2}$. As a consequence, by Theorem \ref{thm: LD2converge}, there exists a small enough $\alpha>0$ such that the sequence $\{Z_k\}$ generated by the PGD converges to $X$ in a linear rate: $\|Z_k-X\|_F\leq e^{-ck}$, with $c=-\log\left(1-\Omega\left(\frac{\alpha}{C_2^2C_3^2}\right)\right)$.
\end{theorem}

By throwing away a controllable failure probability, many random sensing applications potentially have such distance-preserving property. Some examples are mentioned in Section \ref{sec: intro}. The RIP (Restricted Isometry Property) can also be seen as a special case of this distance-preserving condition. Instead of checking the descent inequality and the {\L}ojasiewicz inequality in Theorem \ref{thm: LD2converge}, we use the above conditions as a more user-friendly version for such distance-preserving cases. 

We are now ready to prove Theorems \ref{thm: rank1thm} and \ref{thm: pr}, using Theorem \ref{thm: Converge12} and following a similar but simpler strategy as the one in Section \ref{sec: prooftech}.

\subsection{Proof of Theorem \ref{thm: rank1thm}}\label{sec: pfthm2.2}
\begin{proof}
Denote $Z_+=\phi_{\alpha}(Z)$ and $Z_+=z_+z_+^{\dag}$. Recall that $Z=zz^{\dag}$ and $X=xx^{\dag}$, $h=\|Z\|_F$ and $\rho=\frac{\langle X,Z\rangle}{\|X\|_F\|Z\|_F}$. Let $u_z=\frac{z}{\|z\|}$ and $u_{z,+}=\frac{z_+}{\|z_+\|}$. Then $\sqrt{\rho}=\langle u_z,x\rangle$. By Lemma \ref{lem: dylowrank}, we have 
\begin{align*}
    \frac{d}{dt}h&=u_z^{\dag}(X-Z)u_z=\rho-h,\\
    \frac{d}{dt}u_z&=\frac{1}{h}\cdot(I-u_zu_z^{\dag})(X-Z)u_z =\frac{1}{h}\cdot (\sqrt{\rho}x-\rho u_z),\\
    \frac{d}{dt}\rho &= \frac{2}{h}\cdot \rho (1-\rho).
\end{align*}

Assume $\delta>0$ is a small constant. Since $r=1$, by Lemma \ref{lem: balls}, the only spurious region is $\mathcal{B}(0,\delta)$. Since $Z_0$ is drawn from the general random distribution, with high probability no less than $1-\frac{1}{\poly(n)}$, we have $\rho|_{t=0}\geq \frac{1}{\poly(n)}$. Observe from the third equation above that $\rho$ is non-decreasing, and $\frac{d}{dt}\rho\gtrsim \rho$ until $\rho$ approaches 1. Thus we have $\rho\gtrsim \delta=\Omega(1)$ within $\mathcal{O}(\log n)$ time. As $\rho$ is non-decreasing, the Riemannian gradient flow arrives in $\M\setminus \mathcal{B}(0,\delta)$ and remains there. By Theorem \ref{thm: Converge12} and Lemma \ref{lem: F12singlestep}, it further takes no more than $\mathcal{O}(\log\frac{1}{\epsilon})$ time to generate an $\epsilon$-accuracy solution. Combining all the above, to generate an $\epsilon$-accurate solution, i.e. $\|Z_k-X\|_F\leq \epsilon\|X\|_F$, it takes $\mathcal{O}(\log\frac{1}{\epsilon}+\log n)$ time for the gradient flow.

For the Riemannian gradient descent, by {{Assumption 1}}, we have
\begin{align*}
    h_{k+1}&=h_k+{(\alpha+o(\alpha))}\cdot (-h_k+\rho_k),\\
    \rho_{k+1}&=\rho_k+2{(\alpha+o(\alpha))}\cdot \frac{\rho_k}{h_k}(1-\rho_k).
\end{align*}
Using an argument similar to the continuous case, we can prove it only takes $\mathcal{O}(\log\frac{1}{\epsilon}+\log n)$ iterations to generate an $\epsilon$-accurate solution, i.e. $\|Z_k-X\|_F\leq \epsilon\|X\|_F$.
\end{proof}

\subsection{Proof of Theorem \ref{thm: pr}}\label{sec: pfthm2.3}
\begin{proof}
Recall that $Z=zz^{\dag}$ and $X=xx^{\dag}$, $h=\|Z\|_F$ and $\rho=\frac{\langle X,Z\rangle}{\|X\|_F\|Z\|_F} \in [0,1]$.
If $\F=\R$, the population loss of (\ref{eq: prloss}) is
    \begin{align*}
        \mathbb{E}f(Z)&=\frac{3}{2}\|Z\|_F^2+\frac{3}{2}\|X\|_F^2-\|Z\|_F\|X\|_F-2\langle Z,X\rangle\\
        &=\frac{1}{2}\left(\|Z\|_F-\|X\|_F\right)^2+\|Z-X\|_F^2.
    \end{align*}
    Since $0 \le \left(\|Z\|_F-\|X\|_F\right)^2 \le \|Z-X\|_F^2$, we have 
    $$\|Z-X\|_F^2\leq F(Z)\leq\frac{3}{2}\|Z-X\|_F^2.$$ 
    If $\F=\C$, the population loss of (\ref{eq: prloss}) is
    \begin{align*}
        \mathbb{E}f(Z)&=\|Z\|_F^2+\|X\|_F^2-\|Z\|_F\|X\|_F-\langle Z,X\rangle\\
        &=\frac{1}{2}\left(\|Z\|_F-\|X\|_F\right)^2+\frac{1}{2}\|Z-X\|_F^2.
    \end{align*}
And $\frac{1}{2}\|Z-X\|_F^2\leq F_2(Z)\leq \|Z-X\|_F^2$.

We still denote $Z_+=\phi_{\alpha}(Z)$. Assume $Z=zz^{\dag}$, and $Z_+=z_+z_+^{\dag}$. Let $u_z=\frac{z}{\|z\|}$ and $u_{z,+}=\frac{z_+}{\|z_+\|}$. By Lemma \ref{lem: dylowrank}, we have 
\begin{align}
    \frac{d}{dt}h&=u_z^{\dag}\left(2X-\left(\theta+2-\frac{\theta}{h}\right)Z\right)u_z=2\rho-(2+\theta)h+\theta,\nonumber\\
    \frac{d}{dt}u_z&=(I-u_zu_z^{\dag})\left(2X-\left(\theta+2-\frac{\theta}{h}\right)Z\right)u_z\cdot \frac{1}{h}=\frac{2}{h}\cdot (\sqrt{\rho}x-\rho u_z),\nonumber\\
    \frac{d}{dt}\rho &= \frac{4}{h}\cdot \rho (1-\rho). \label{eq: rhody}
\end{align}

Note that in this case, in addition to the spurious region $\mathcal{B}(0,\delta)$, there is another region $\mathcal{B}(Z^*,\delta):=\{Z:\rho\lesssim \delta,|\theta+2-\frac{\theta}{h}|\lesssim \delta\}$ where the Riemannian gradient is $\delta$-small, because the Riemannian gradient is now
\begin{align*}
    \grad_{\mathcal{M}}F_2(Z)=\left(\theta+2-\frac{\theta}{h}\right)Z-2P_{T_z}(X).
\end{align*}
Similar to Lemma \ref{lem: balls}, we have
\begin{equation}\label{eq: 2.3sncon}
    \|P_{T_z}\left(\grad_{\mathcal{M}}F_2(Z)\right)\|\leq\delta \iff Z\in\mathcal{B}(0,\delta)\cup \mathcal{B}(Z^*,\delta)\Rightarrow \rho\lesssim \delta.
\end{equation}

Since $Z_0$ is drawn from the general random distribution, with high probability no less than $1-\frac{1}{\poly(n)}$, we have $\rho|_{t=0}\geq \frac{1}{\poly(n)}$. Observe from (\ref{eq: rhody}) that $\frac{d}{dt}\rho\gtrsim \rho$ when $\rho<\frac{1}{2}$ and $h$ is bounded. The boundedness of $h$ can be easily concluded from the first equation using $0\le\rho\le 1$ and $0<\theta < \Omega(1)$. Thus we have $\rho\gtrsim \delta=\Omega(1)$ within $\mathcal{O}(\log n)$ time. Using the non-decreasing property of $\rho$ and the relation (\ref{eq: 2.3sncon}), we conclude that the Riemannian gradient flow arrives in $\M\setminus \mathcal{B}(0,\delta)\cup\mathcal{B}(Z^*,\delta)$ and remains there. By Theorem \ref{thm: Converge12} and Lemma \ref{lem: F12singlestep}, it further takes no more than $\mathcal{O}(\log\frac{1}{\epsilon})$ time to generate an $\epsilon$-accuracy solution. Combining all the above, to generate an $\epsilon$-accurate solution, i.e. $\|Z_k-X\|_F\leq \epsilon\|X\|_F$, it needs $\mathcal{O}(\log\frac{1}{\epsilon}+\log n)$ time.

For the Riemannian gradient descent, by {{Assumption 1}}, we have
\begin{align*}
    h_{k+1}&=h_k+(\alpha+o(\alpha))\cdot (\theta-(2+\theta)h_k+2\rho_k),\\
    \rho_{k+1}&=\rho_k+4(\alpha+o(\alpha)) \cdot \frac{\rho_k}{h_k}(1-\rho_k).
\end{align*}
Similar to the argument for continuous case, we can prove it only takes $\mathcal{O}(\log\frac{1}{\epsilon}+\log n)$ iterations to generate an $\epsilon$-accurate solution, i.e. $\|Z_k-X\|_F\leq \epsilon\|X\|_F$.

\end{proof}

\subsection{Comparison of Theorem \ref{thm: rank1thm} and Theorem \ref{thm: pr}}\label{sec: connection23}
\noindent{\bf{Dynamics. }}The dynamical low-rank approximation (Lemma \ref{lem: dylowrank}) shows that the evolution of the column space is given by $$\dot{U}_z=P_{U_z}^{\perp}\left(-\grad F(Z)\right)U_zS^{-1}.$$
For $F_1(Z)=\frac{1}{2}\|Z-X\|_F^2$, we have that $\grad F_1(Z)=Z-X$, while for $F_2(Z)=\frac{\theta}{2}(\|Z\|_F-\|X\|_F)^2+\|Z-X\|_F^2$ we have $\grad F_2(Z)=(2+\theta-\frac{\theta}{h})Z-2X$. Although $F_2(Z)$ only satisfies the weak isometry property, direct computation shows that $\dot{U}_z$ for $F_1(Z)$ and $F_2(Z)$ are similar, because $P_{U_z}^{\perp}$ on the right cancels out the $Z$ terms and leaves only the $X$ terms. That explains why the dynamics of $F_1(Z)$ is similar to that of $F_2(Z)$.

\smallskip
\noindent{\bf{Stationary points.}} Theorem \ref{thm: rank1thm} is a special case of Theorem \ref{thm: rankrconverge}, therefore $Z^*=0$ is the only spurious critical point and has a saddle-like property (see Section \ref{sec: geometry_sr}). On the other hand, for the phase retrieval problem in Theorem \ref{thm: pr}, it has two groups spurious critical points, which are $Z^*=0$ and $\{Z^*: \|Z^*\|_F=\frac{\theta}{2+\theta}\|X\|_F,\langle Z,X\rangle =0\}$. Still, the upper bound for the number of iterations that the sequence is trapped by the spurious region can be estimated in a similar way. This can be seen by comparing the proofs in Section \ref{sec: pfthm2.2} and Section \ref{sec: pfthm2.3}.

\begin{figure}[ht]
\centering
\begin{subfigure}[t]{0.30\textwidth}
\centering
\includegraphics[width=\linewidth]{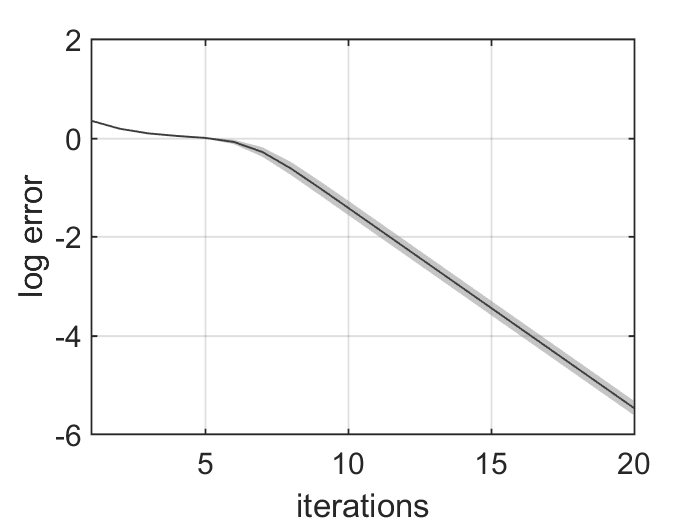}
\caption{\emph {Log-error}}\label{fig: rank1_logerror}
\end{subfigure}
\hspace{0.03\textwidth}
\begin{subfigure}[t]{0.30\textwidth}
\centering
\includegraphics[width=\linewidth]{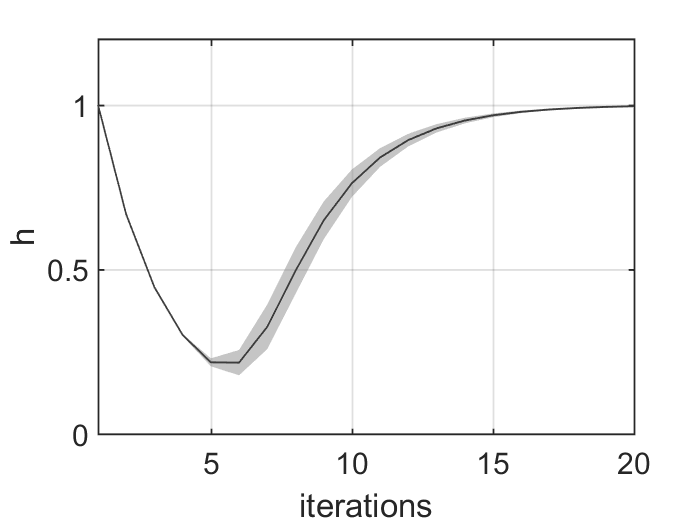}
\caption{\emph{The evolution of $h$}}\label{fig: rank1_h}
\end{subfigure}
\hspace{0.03\textwidth}
\begin{subfigure}[t]{0.30\textwidth}
\centering
\includegraphics[width=\linewidth]{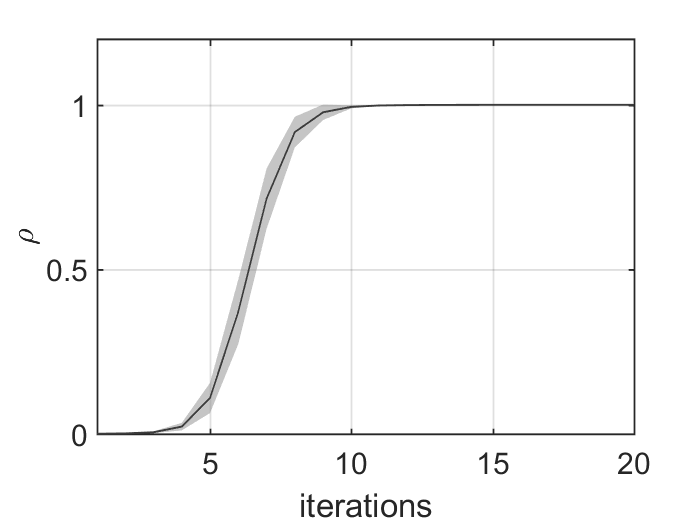}
\caption{\emph{The evolution of $\rho$}}\label{fig: rank1_rho}
\end{subfigure}
\caption{
Solving the rank-1 matrix recovery by the randomly initialized PGD, with $n=1024$, $\alpha=\frac{1}{3}$. Each band stands for the results from 100 independent experiments. }
\label{fig: rank1}
\end{figure}

\begin{figure}[ht]
\centering
\begin{subfigure}[t]{0.30\textwidth}
\centering
\includegraphics[width=\linewidth]{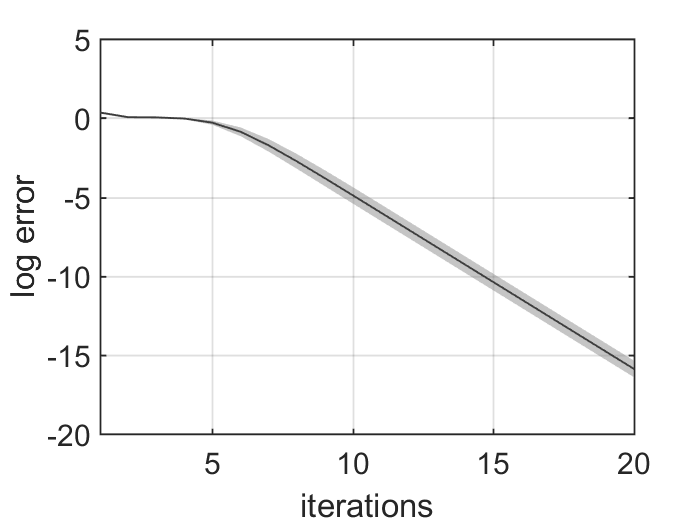}
\caption{\emph {Log-error}}\label{fig: pr_logerror}
\end{subfigure}
\hspace{0.03\textwidth}
\begin{subfigure}[t]{0.30\textwidth}
\centering
\includegraphics[width=\linewidth]{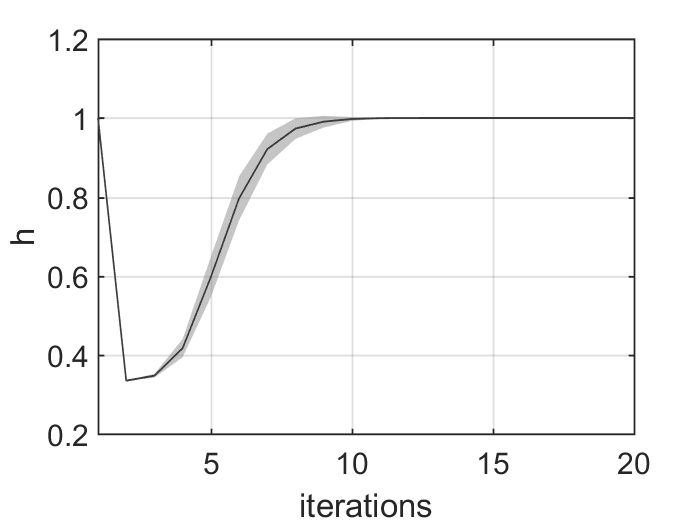}
\caption{\emph{The evolution of $h$}}\label{fig: pr_h}
\end{subfigure}
\hspace{0.03\textwidth}
\begin{subfigure}[t]{0.30\textwidth}
\centering
\includegraphics[width=\linewidth]{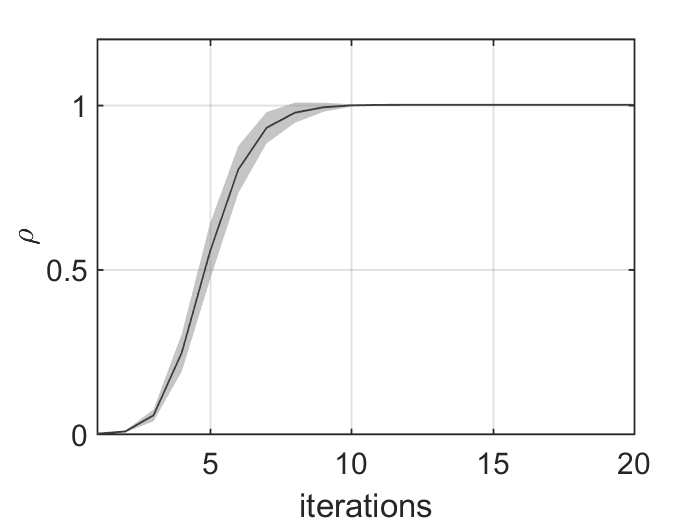}
\caption{\emph{The evolution of $\rho$}}\label{fig: pr_rho}
\end{subfigure}
\caption{
Solving the population phase retrieval problem by the randomly initialized PGD, with $n=1024$, $\alpha=\frac{1}{3}$. Each band stands for the results from 100 independent experiments. }
\label{fig: pr}
\end{figure}

\smallskip
\noindent{\bf{Numerical illustration.}} To see the similarity between the evolution behavior in solving the rank-1 matrix recovery and the phase retrieval problem, we give some numerical experiments in Figure \ref{fig: rank1} and Figure \ref{fig: pr} for a comparison. We can see that the curves of the evolution of $h$ and $\rho$ have similar shapes in both problems.

\section{Conclusion and future work}\label{sec: Conclusion}
In this paper, we have established a unified framework for the analysis of a class of low-rank matrix recovery problems. We have shown that using the Riemannian gradient descent (the PGD) algorithm on the low-rank matrix manifold, there is rigorous theoretical guarantee for the fast convergence rate in low-rank matrix recovery problems. 

For this purpose, we first performed an extensive analysis of the low-rank matrix manifold $\M_r$ itself by analyzing the simple least squares loss function $F_1(Z) = \frac{1}{2}\|Z-X\|_F^2$ where $X$ is the ground truth. Our focus is on the symmetric positive semi-definite (SPSD) setting which is common in practice. Our results on the rank-r manifold with $r>1$ are original and they are much more complicated than the corresponding results for the rank-1 case.

We showed that there is a ground truth and several spurious critical points on the manifold. The spurious critical points are of independent interest themselves, as they behave like strict saddle points, but their Hessians have singular eigen directions. 
We proved that the gradient descent or gradient flow starting from an initial guess drawn from the general random distribution converges to the ground truth with high probability. 
The initializations that might lead to the spurious critical points only have a small probability measure on the manifold. 
Improvement of this result to zero measure is left for future work.

The convergence rate towards the ground truth is nearly linear and is essentially independent of the dimensionality of the problem. The major difficulty when analyzing the convergence rate comes from estimating the upper bound for the number of iterations that the sequence is trapped by the spurious regions. Our primary tool is the iteration function of the column space derived from the dynamical low-rank approximation. 
We showed that with high probability, the initial angle between the column spaces of the ground truth and the random initialization point is $\widetilde{\Omega}(\frac{1}{n})$, i.e. $\Omega(\frac{1}{n})$ with possible additional log-factors. The angle grows fast in spurious regions. Thus, we showed that with high probability, the sequence generated by the randomly initialized PGD escapes from the spurious regions quickly and enters a good region. When the sequence enters the good region, we then used the {\L}ojasiewicz inequality tool to derive linear convergence.

The above analysis offers a general framework for a class of inverse problems that share a desirable structure, namely those problems whose forward problem is a linear mapping from a low-rank matrix to a vector and preserves the isometry property to some extent. The well-known RIP ensemble is a special case, but there are other applications with a weak isometry property.
We analyzed the phase retrieval problem as an example of weak isometry problems, and established nearly optimal (linear) convergence rate.
We focused on the population problem, i.e. the expectation of the loss function with respect to sampling. We invoked its connection with the rank-1 simple least squares problem, which can be described by scalar ODEs instead of matrix ODEs.

The global analysis for population loss functions can also be extended to finite-sample problems. The finite-sample loss function is the sum of its population loss plus some small deviations. One can control the magnitude of the deviations and show that the loss function still satisfies some weak isometry conditions. Thus the fundamental convergence guarantee by the \L ojasiewicz inequality is readily applicable on the majority of the manifold. On the other hand, the geometry of the spurious regions, as well as the escape from these spurious regions, could be different from the population case. We leave the detailed analysis of finite-sample cases to our future work.

\vspace{8pt}

{\textbf{Acknowledgements.}} This research was in part supported by NSF Grants DMS P2259068 and DMS P2259075. We would also like to thank Prof. Jian-feng Cai for helpful suggestions.

\bibliographystyle{plain}
\bibliography{reference}
\appendix

\section{Manifold setting and optimization algorithm}\label{sec: pre}

\subsection{Building blocks: low-rank matrix manifold} 
In this subsection we first present some basic structural properties of the low-rank matrix manifold as a preliminary. As mentioned in our previous work \cite{paper1}, the usual fixed-rank manifold may not be rigorous enough for two reasons stated below:
\begin{enumerate}[label=\arabic*)]
    \item The fixed-rank manifold $\M_r$ is not closed. Iterative optimization techniques can  generate a sequence towards the ground truth, and closedness is naturally necessary for asymptotic convergence analysis.
    \item  It is possible that at some step $(Z_k - \alpha P_{T_{Z_k}} (\grad f(Z_k)))$ happens to have rank lower than $r$ and falls outside $\M_r$. 
\end{enumerate}  
Therefore, for theoretical completeness we need the notion $\overline{\M_r}=\{Z\in\F^{n_1\times n_2}: \text{rank}(Z)\leq r\}$ instead. The following are some basic definitions and essential properties on $\overline{\M_r}$. 

\begin{definition}[Tangent space]
Let $Z\in\M_r$, $Z=U\Sigma V^{\dag}$. Denote $\mathcal{U}=\text{Col}(U)$, $\mathcal{V}=\text{Col}(V)$ as the column spaces of $U$ and $V$ respectively. Then the \emph{tangent space} of $\M_r$ at $Z$ is 
    \begin{equation*}
        T_Z\M_r=(\mathcal{U}\otimes\mathcal{V})\oplus(\mathcal{U}\otimes\mathcal{V}^\perp)\oplus (\mathcal{U}^\perp\otimes\mathcal{V}).
    \end{equation*}
    The projection onto the tangent space is
    \begin{equation*}
        P_{T_z}=P_{\mathcal{U}}\otimes I + I\otimes P_{\mathcal{V}}- P_{\mathcal{U}}\otimes P_{\mathcal{V}}.
    \end{equation*}
\end{definition}

The metric on $T\M$ is inherited from the metric of the embedded Euclidean space $\F^{n_1\times n_2}$ which is equipped with inner product defined as $\langle A,B\rangle=\text{tr}(A^{\dag}B)$ and Frobenius norm $\|A\|_F=\sqrt{\text{tr}(A^{\dag}A)}$.

\begin{definition}[Tangent cone]
     Let $Z\in\M_s\subset(\overline{\M_r}\backslash \M_r)$ where $s<r$,  $Z=U\Sigma V^{\dag}$, $\mathcal{U}=\text{Col}(U)$, $\mathcal{V}=\text{Col}(V)$. Then the \emph{tangent cone} of $\overline{M_r}$ at $Z$ is defined as follows
    \begin{equation*}
        T_Z\overline{M_r}=T_Z\M_s\oplus\{\eta: \eta\in \mathcal{U}^\perp\otimes\mathcal{V}^\perp\mbox{, and }\text{rank}(\eta)=r-s\}
    \end{equation*}
    The projection onto the tangent cone is given by
    \begin{align*}
     P_{T_z\overline{\M_r}}(Y) = P_{T_z\M_s}(Y) + Y_{r-s}
    \end{align*}
    where $Y_{r-s}$ is the best rank ($r-s$) approximation of $Y-P_{T_Z\M_s}(Y)$ in the Frobenius norm.
\end{definition}
We use $P_{T_z}$ for both $P_{T_z\M_r}$ and $P_{T_z\overline{\M_r}}$ when there is no confusion. 

\begin{remark}
    The low-rank matrix manifold as presented above is for general non-Hermitian case. The Hermitian case is similar except that the fixed rank manifold may have branches.
\end{remark}

\subsection{Optimization technique: the projected gradient descent (PGD)}\label{sec: PGD}
In this section, we introduce the optimization technique on the low-rank matrix manifold, namely the projected gradient descent (PGD) with soft retraction onto the manifold. This Riemannian gradient descent technique has been studied in \cite{Reinhold,ManiOpt1,ManiOpt2,cai2018solving}. For example, \cite{ManiOpt2} and \cite{cai2018solving} use the Riemannian gradient descent to solve low-rank matrix recovery problems. These works also point out that the PGD enjoys light computational cost.
In this paper, we will focus on the global analysis of such manifold optimization technique with random initialization.

Assume we are given a differentiable objective function $f(\cdot):\M\rightarrow \R$ to be minimized, where $\M$ can be a general Riemannian manifold. We start from a random  initial guess $Z_0\in\M$. Assume the sequence $\{Z_k\}_{k=0}^K$ is generated by the PGD (projected gradient descent):
\begin{equation}\label{eq: algorithm}
    Z_{k+1}=\mathcal{R}\left(Z_k - \alpha_k P_{T_{Z_k}}(\grad f(Z_k))\right).
\end{equation}
Here $P_{T_{Z_k}}$ is the projection onto the tangent space (or tangent cone at a rank-deficient point) of $\M$ at point $Z_k$, $\alpha_k$ is the $k$-th stepsize, and $\mathcal{R}: T_{Z} \rightarrow \M$ is a retraction defined in definition \ref{def: retraction}. 

Specifically, for the loss function $f$ as in (\ref{eq: maniopt}), we have the following algorithm: \\
$\rule[-2pt]{16.5cm}{0.1em}\\
\mbox{\bf{Projected Gradient Descent (PGD)} } \\
\rule[4pt]{16.5cm}{0.05em}\\
\mbox{{\bf{Input: }}Random measurements } {\{A_j\}}_{j=1}^{m} \mbox{, information } {\{y_j\}}_{j=1}^{m},   \mbox{random initialization $Z_0$,}\\ \mbox{maximal iteration number $K_0$, tolerance $\epsilon$.}\\ \\
\mbox{{\bf{Iteration: }} For $k=1:K_0$, do:}$
$$Z_{k+1}=\mathcal{R}\left(Z_k-P_{T_{Z_k}}\left(\frac{1}{m }\sum_{j=1}^m\langle Z_k-X,A_j\rangle A_j\right)\right),\mbox{ if $f(Z_k)<\epsilon$, break.}$$
$\mbox{\bf{Output: }}  \mbox{Estimator } \hat{X}=Z_k.\\
\rule[-2pt]{16.5cm}{0.1em}$\\

In the aforementioned algorithm, the retraction operation is necessary since it ensures the generated iteration points stay on the manifold $\M$. The retraction operator is defined as follows.

\begin{definition}[Retraction] \label{def: retraction}
Let $T_{Z}$ be the tangent space (or tangent cone) of $\M$ at $Z$. We call $\mathcal{R}_{Z}:T_{Z}\rightarrow\M$ a {\emph{retraction}}, if for any $\xi\in T_{Z}$,
    \begin{align}\label{eq: first_order_retraction}
        \lim_{\alpha\rightarrow 0^+}\frac{\|\mathcal{R}_{Z}(\alpha\xi)-(Z+\alpha\xi)\|_F}{\alpha}=0.
    \end{align}
For simplicity we may also write $\mathcal{R}_{Z}(\alpha\xi)$ as $\mathcal{R}(Z+\alpha\xi)$. We will refer to (\ref{eq: first_order_retraction}) as the first-order retraction property. 
\end{definition}

\begin{remark}
    A natural retraction on $\M_r$ is the following best rank-$r$ approximation under the Frobenius norm:
    \begin{align*}
        \mathcal{R}_N(Z+\xi) = \mathop{\arg\min}_{Y \in \M_r} \|Z+\xi-Y\|_F .
    \end{align*}
    We use $\mathcal{R}_N$ as our $\mathcal{R}$ in the rest of the paper. It not only satisfies the first-order retraction property, but is also second order, see also \cite{FirstRetract,Reinhold,ManiOpt1}:
    \begin{align*}
        \mathcal{R}(Z+\alpha\xi) = Z+\alpha\xi + \alpha^2\eta+\mathcal{O}(\alpha^3).
    \end{align*}
\end{remark}

It is worth mentioning that there exists other manifold optimization techniques. For example, some manifold optimization methods skip the projection step $P_{T_{Z_k}}$ and compute the singular value decomposition directly. This is also called the ``hard retraction''. We choose the current projected gradient descent with ``soft retraction'' mainly due to two reasons:
\begin{enumerate}[label=\arabic*)]
    \item Most of the operations we list are defined only for tangent bundles, e.g. first-order retraction (\ref{eq: first_order_retraction}) and the Riemannian Hessian. Generally, $\grad f(Z_k)$ is not in $T_{Z_k}$, but $P_{T_{Z_k}}(\grad f(Z_k))$ is. 
    \item The projected version of GD is also cheaper in terms of computation. Namely, solving $\mathcal{R}(Z_k - \alpha_k \grad f(Z_k))$ involves calculating SVD of a $n_1\times n_2$ matrix, while $\mathcal{R}(Z_k - \alpha_k P_{T_{Z_k}}(\grad f(Z_k)))$ only involves that of a $2r\times 2r$ matrix, as mentioned in the previous literature \cite{ManiOpt2}.
    Since $r\ll \min\{n_1,n_2\}$, the soft retraction is of lighter computational cost.
\end{enumerate}

\noindent
\textbf{Light computational cost of the PGD.} With the PGD, we have
    \begin{align*}
        Z_{k+1}&=\mathcal{R}(Z_k-\alpha P_{T_{Z_k}}(W_k))\\
        &=\mathcal{R}(U_k\Sigma_kV_k^\dag-\alpha \left( U_k U_k^\dag W_k+W_k V_kV_k^\dag-U_kU_k^\dag W_k V_kV_k^\dag\right))\\
        &=\mathcal{R}(U_k\left(\Sigma_k-\alpha U_k^\dag W_k V_k\right)V_k^\dag-\alpha U_k((I-V_kV_k^\dag)W_k^\dag U_k)^\dag\\
        &\quad-\alpha((I-U_kU_k^\dag)W_kV_k)V_k^\dag)\\
        &=\mathcal{R}\left(\begin{pmatrix}
        U_k &Q_{k,2}\\
        \end{pmatrix}
        \begin{pmatrix}
        \Sigma_k-\alpha U_k^\dag W_k V_k &-\alpha R_{k,1}^\dag \\
        -\alpha R_{k,2}&0 \\
        \end{pmatrix}
        \begin{pmatrix}
        V_{k,1}^\dag\\
        Q_{k,1}^\dag
        \end{pmatrix}\right).
    \end{align*}
    Here $W_k=\grad f(Z_k)\in\F^{n_1\times n_2}$  and we  assume $Z_k=U_k\Sigma_k V_k^\dag$. Assume $(I-V_kV_k^\dag)W_k^\dag U_k=Q_{k,1}R_{k,1}$ and $(I-U_kU_k^\dag)W_kV_k=Q_{k,2}R_{k,2}$ are the QR factorizations of the respective matrices. Notice that $Q_{k,2}^\dag U_k=\mathbf{0}_{r\times r}$, $Q_{k,1}^\dag V_k=\mathbf{0}_{r\times r}$. Therefore, to compute SVD of $Z_k-\alpha P_{T_{Z_k}}(W_k)$ it only involves solving the SVD of $
        \begin{pmatrix}
        \Sigma_k-\alpha U_k^\top W_k V_k &-\alpha R_{k,1}^\dag \\
        -\alpha R_{k,2}&0 \\
        \end{pmatrix}
    $, which is only a $2r\times 2r$ matrix.

\section{Proofs}\label{sec: Proofs}
\subsection{Proof of Theorem \ref{thm: LD2converge}}
\begin{proof}
Since $\{f_k\}_{k=1}^{\infty}$ is a monotone and lower bounded sequence, Condition (\ref{eq: D}) and continuity of $f(.)$ implies convergence of $\{Z_k\}$ to some fixed point $Z^*$.
Without loss of generality we assume that $f(Z^*)=0$. By Conditions (\ref{eq: D}) and (\ref{eq: L}), we have
\begin{align*}
&\|Z_{k+1}-Z_k\|_F\leq \frac{1}{C_d \|P_{T_z}(\grad f(Z_k))\|_F}(f_k-f_{k+1})\leq \frac{C_l}{C_d}f_k^{\omega-1}(f_k-f_{k+1})\\
& \leq \frac{C_l}{C_d} \int_{f_{k+1}}^{f_k} \phi^{\omega-1} d\phi=\frac{C_l}{\omega C_d}(f_k^{\omega}-f_{k+1}^{\omega}).
\end{align*}
Since $\{f_k\}_{k=1}^{\infty}$ is a monotone and lower bounded sequence, $\{f_k\}$ is convergent. Therefore, $\{Z_k\}$ is convergent, and the limit point is 
$Z^*$.

Consider $s_k:=\sum_{i=k}^{\infty}\|Z_{i+1}-Z_i\|_F\leq \frac{C_l}{\omega C_d}f_k^{\omega}$. Then we get
\begin{align*}
    s_k^{\frac{1-\omega}{\omega}}\leq (\frac{C_l}{\omega C_d})^{\frac{1-\omega}{\omega}}f_k^{1-\omega}\leq C_l(\frac{C_l}{\omega C_d})^{\frac{1-\omega}{\omega}}\|P_{T_{Z_k}}(\grad f(Z_k))\|_F.
\end{align*}
Let $\xi_k=-P_{T_{Z_k}}(\grad f(Z_k))$ and $\alpha\widetilde{\xi}_k=Z_{k+1}-Z_k=\mathcal{R}(Z_k+\alpha\xi_k)-Z_k$. By the retraction property, $\alpha\widetilde{\xi}_k=\alpha\xi_k+o(\alpha)$. So we have 
\begin{align*}
    s_k^{\frac{1-\omega}{\omega}}&\leq C_l(\frac{C_l}{\omega C_d})^{\frac{1-\omega}{\omega}}\|\xi_k\|_F= C_l(\frac{C_l}{\omega C_d})^{\frac{1-\omega}{\omega}} \frac{1}{\alpha}(\|Z_{k+1}-Z_k\|_F +o(\alpha)).
\end{align*}
Let $\rho=\rho(\alpha):=\frac{\alpha }{C_l}(\frac{\omega C_d}{C_l})^{\frac{1-\omega}{\omega}}$, i.e. $\rho$ is a constant depending only on $C_l$, $C_d$ and $\omega$. Then we have
\begin{align*}
    \rho(\alpha) s_k^{\frac{1-\omega}{\omega}} = \|Z_{k+1}-Z_k\|_F + o(\alpha).
\end{align*}
One can choose $\alpha$ small enough such that $0<\rho(\alpha)<1$ and 
\begin{equation}\label{eq: rhosk}
    \rho(\alpha) s_k^{\frac{1-\omega}{\omega}}\leq s_k-s_{k+1}.
\end{equation}

In the case of $\omega=\frac{1}{2}$, the above inequality gives $s_{k+1}\leq (1-\rho)s_k$, which implies $\|Z_k-Z^*\|_F\leq s_k\leq (1-\rho)^k s_0=s_0e^{-ck}$, where $c=-\log(1-\rho)=-\log(1-\frac{\alpha C_d}{2 C_l^2})>0$. This gives the linear convergence rate.

In the case of $0<\omega<\frac{1}{2}$: Assume $s_k=c_1 k^{-p}$, then we have
    \begin{align*}
        s_{k+1}&=c_1\frac{1}{(k+1)^p}=c_1\frac{1}{k^p}(1+\frac{1}{k})^{-p}\\
        & = c_1\frac{1}{k^p}(1-\frac{p}{k})+O(\frac{1}{k^{p+2}})\\
        &=s_k(1-p s_k^{\frac{1}{p}} c_1^{-\frac{1}{p}})+O(\frac{1}{k^{p+2}}) \\
        &\leq s_k(1-\rho s_k^{\frac{1-2\omega}{\omega}}),
    \end{align*}
    where the last inequality follows equation \eqref{eq: rhosk}. Choose $c_1$ large enough, then the above inequality holds with $p = \frac{\omega}{1-2\omega}$. Thus, we obtain $\|Z_k-Z^*\|_F\leq s_k\leq c_1 k^{-\frac{\omega}{1-2\omega}}$. This implies the polynomial convergence rate.
\end{proof}

\subsection{Proof of Lemma \ref{lem: F12singlestep}}
\begin{proof}
For any finite $k\in [K]$, from condition (D) and (L), by the first-order retraction property, we have:
\begin{align*}
    f_{k+1}&\leq f_k-C_d\|P_{T_{Z_k}}(\grad f(Z_k))\|\|Z_{k+1}-Z_k\|\\
    &\leq f_k-\Omega(\alpha)C_d\|P_{T_{Z_k}}(\grad f(Z_k))\|^2\\
    & \leq \left(1-\Omega(\alpha)\frac{C_d}{C_l^2}\right)f_k.
\end{align*}
Since for both $F_1(Z)$ and $F_2(Z)$, we have $\|Z_k-X\|_F^2\lesssim f_k$, we have $\|Z_k-X\|_F\lesssim e^{-ck}$, with $\rho=\Omega(\alpha\frac{C_d}{C_l^2})$ and $c=-\frac{1}{2}\log(1-\rho)$. If $\alpha>0$ is properly small, we have $c>0$.
\end{proof}

\subsection{Proof of Lemma \ref{lem: spurious_fp}}
\begin{proof}
\begin{enumerate}[label=\arabic*)]
    \item    It is obvious that $Z$ is a fixed point of $Z_{k+1}=\mathcal{R}(Z_k-\alpha P_{Z_k}(\grad f(Z_k)))$ if and only if $P_{T_z}(\grad F_1(Z))=0$. Denote $Z= U_z \Sigma_z V_z^{\dag}$ and $X=UDV^{\dag}$, then for any $\xi \in T_{Z}$, there exists $\Delta_1\in\F^{n_1\times r}$ and $\Delta_2\in\F^{n_2\times r}$, such that $\xi=U_z\Delta_1^{\dag}+\Delta_2V_z^{\dag}$. Simple calculation gives
    \begin{align*}
        & P_{T_z}(\grad F_1(Z))=0
        \Longleftrightarrow \langle P_{T_z}(Z-X), \xi\rangle = 0\text{, for all }\xi\in T_Z \\
        \Longleftrightarrow & \langle Z-X, U_z\Delta_1^{\dag}+\Delta_2V_z^{\dag}\rangle = 0 \text{, for all }\xi=U_z\Delta_1^{\dag}+\Delta_2V_z^{\dag}\in T_Z\\
        \Longleftrightarrow &  \text{tr}((V_z\Sigma_z-VD U^{\dag}U_z)\Delta_1^{\dag}+(\Sigma_zU_z^{\dag}-V_z^{\dag}VD U^{\dag})\Delta_2)\text{, for all }\Delta_1\in\F^{n_1\times r}, \Delta_2\in\F^{n_2\times r}\\
        \Longleftrightarrow & V_z\Sigma_z-VD U^{\dag}U_z=\Sigma_zU_z^{\dag}-V_z^{\dag}VD U^{\dag}=0\\
        \Longleftrightarrow &
        U_z^{\dag}X=\Sigma_z V_z^{\dag} \text{ and }U_z\Sigma_z=XV_z.
    \end{align*}    
    This implies $ P_{U_z}(X)=P_{U_z}(Z)=Z\text{ and }P_{V_z}(Z^{\dag})=Z^{\dag}=P_{V_z}(X^{\dag})$. Assume 
    \begin{align*}
        X=\begin{pmatrix}U_z & \widetilde{U}_z
        \end{pmatrix} \begin{pmatrix} X_{11}&X_{12}\\ 
        X_{21} &X_{22}
        \end{pmatrix}
        \begin{pmatrix}
        V_z^{\dag}\\
        \widetilde{V}_z^{\dag}
        \end{pmatrix}.
    \end{align*}
    Then we get $X_{11}=\Sigma_z$, $X_{12}=0$ and $X_{21}=0$. Therefore, we have
    $$Z=U_1D_1V_1^{\dag}\text{, with }U =\begin{pmatrix} U_1 &U_2\end{pmatrix}\text{, } D = \text{diag } \{D_1, D_2\} \text{ and }  V = \begin{pmatrix} V_1 &V_2\end{pmatrix}.$$
    \item If $X$ has distinct singular values, then $\mathcal{S}$ consists of the points $Z^*=\sum_{i=1}^r d_i\eta_i u_i v_i^{\dag}$, where $\eta\in\{0,1\}^r$ and $\eta$ is not $(1,1,...,1)^{\dag}\in\R^r$. So $|\mathcal{S}|=2^r-1$.
\end{enumerate}
\end{proof}

\subsection{Proof of Lemma \ref{lem: balls}}
\begin{proof}
    Assume $Z=U_z\Sigma_zV_z^{\dag}$, and let $\widetilde{U}_z$, $\widetilde{V}_z \in \F^{n\times (n-r)}$ be the orthogonal complements of $U_z$ and $V_z$. We can express $X$ in the following block form under this new basis:
    \begin{align*}
        X  &= \begin{pmatrix}U_z& \widetilde{U}_z\end{pmatrix}
        \widetilde{X} \begin{pmatrix}
            V_z^{\dag} \\
            \widetilde{V}_z^{\dag}
        \end{pmatrix}
        =   \begin{pmatrix}U_z & \widetilde{U}_z \end{pmatrix}
        \begin{pmatrix}
            X_{11} & X_{12} \\
            X_{21} & X_{22}
        \end{pmatrix} 
        \begin{pmatrix}
            V_z^\dag \\
            \widetilde{V}_z^\dag
        \end{pmatrix},
    \end{align*}
    where
    \begin{align*}
        \widetilde{X} &= \begin{pmatrix}
            X_{11} & X_{12} \\
            X_{21} & X_{22}
        \end{pmatrix} = Q_L \cdot D \cdot Q_R^\dag.
    \end{align*}
    Then we have
    \begin{align*}
        P_{T_z}(X) = \begin{pmatrix}U_z & \widetilde{U}_z \end{pmatrix}
        \begin{pmatrix}
            X_{11} & X_{12} \\
            X_{21} & 0
        \end{pmatrix} 
        \begin{pmatrix}
            V_z^\dag \\
            \widetilde{V}_z^\dag
        \end{pmatrix}, \quad U = \begin{pmatrix}U_z & \widetilde{U}_z \end{pmatrix} Q_L, \quad V = \begin{pmatrix}V_z & \widetilde{V}_z \end{pmatrix} Q_R.
    \end{align*}
    
    Assume that $\|P_{T_z}(Z-X)\|_F \le \delta$, then
	\begin{align*}
		\|P_{T_z}(Z-X)\|_F = \left\|
		\begin{pmatrix}
			X_{11}-\Sigma_z & X_{12} \\
			X_{21} & 0
		\end{pmatrix}\right\|_F = \left\|\widetilde{X} - 
		\begin{pmatrix}
			\Sigma_z & 0 \\
			0 & X_{22}
		\end{pmatrix}\right\|_F \le \delta.
	\end{align*}
	Let
	\begin{align*}
	    \widetilde{S} :=
	    \begin{pmatrix}
			\Sigma_z & 0 \\
			0 & X_{22}
		\end{pmatrix} = 
		\begin{pmatrix}
		    I_r & 0\\
		    0 & P_L
		\end{pmatrix}
		\begin{pmatrix}
		    \Sigma_z & 0
		    \\0 & \Sigma_{22}
		\end{pmatrix}
		\begin{pmatrix}
            I_r & 0
            \\0& P_R^{\dag}
        \end{pmatrix}
	\end{align*}
	where the second equality gives the singular value decomposition of the matrix $\widetilde{S}$. Then $\|\widetilde{X} - \widetilde{Z}\|_F \le \delta$.
	
	Using Lemma \ref{lem: Hoff-W}, we have that the singular values of $\widetilde{S}$ are $\delta$-perturbations of those of $\widetilde{X}$. Note that the singular values of $\widetilde{X}$ are the same as those of $X$, which are $\{d_1,\ldots, d_r\}\cup \{0\}$. On the other hand, the singular values of $\widetilde{S}$ are $\{\sigma_1,\ldots,\sigma_r\} \cup \{\widetilde{\sigma}_1,\ldots,\widetilde{\sigma}_{n-r}\}$, where $\{\sigma_i\}$ and $\{\widetilde{\sigma}_i\}$ are the  diagonal entries of $\Sigma_z$ and $\Sigma_{22}$ respectively. Thus, for each $\sigma_i$, $i \in [r]$, either $|\sigma_i-d_j| = \mathcal{O}(\delta)$ for some $j\in[r]$, or $\sigma_i= \mathcal{O}(\delta)$. In other words, each $\sigma_i$ either captures a singular value of $X$, or is close to zero.
	
	Now let $\mathcal{I}$ denote the set of indices of $d_j$'s captured by $\sigma_i$'s, and $\mathcal{I}^c = [r]\backslash \mathcal{I}$. Without loss of generality, assume $|\sigma_i-d_i| = \mathcal{O}(\delta)$ for $i\in\mathcal{I}$, i.e. their indices also match. Let $U_1 = U(:,\mathcal{I})$, $U_2 = U(:,\mathcal{I}^c)$, $V_1 = V(:,\mathcal{I})$, $V_2 = V(:,\mathcal{I}^c)$, and $D_1 = D(\mathcal{I},\mathcal{I})$, $D_2 = D(\mathcal{I}^c,\mathcal{I}^c)$. Then $\|\Sigma_z(\mathcal{I},\mathcal{I})-D_1\| = \mathcal{O}(\delta)$, and $\|\Sigma_z(\mathcal{I}^c,\mathcal{I}^c)\| = \mathcal{O}(\delta)$. 
	
	By Lemma \ref{lem: sin_theta}, the singular subspaces of $\widetilde{X}$ and $\widetilde{S}$ corresponding to indices $\mathcal{I}$ are $\delta$-close. 
	In mathematical terms, we have
	\begin{align*}
	    &\|P_{I(:,\mathcal{I})} - P_{Q_L(:,\mathcal{I})}\| = \mathcal{O}(\delta), \qquad \|P_{I(:,\mathcal{I})} - P_{Q_R(:,\mathcal{I})}\| = \mathcal{O}(\delta), 
	\end{align*}
	where $I$ denotes the identity matrix and $P$ denotes the projection onto the space spanned by the column vectors. Using the relations $U = (U_z,\widetilde{U}_z) Q_L$, and $V = (V_z, \widetilde{V}_z)Q_R$, we have 
	\begin{align*}
	    &\|P_{U_z(:,\mathcal{I})} - P_{U_1}\| = \mathcal{O}(\delta), \qquad \|P_{V_z(:,\mathcal{I})} - P_{V_1}\| = \mathcal{O}(\delta), 
	\end{align*}
	Let $B= U_z(:,\mathcal{I})$ and $C = V_z(:,\mathcal{I})$ and we have the results regarding $B$ and $C$ in the lemma.
	
    Similarly, the singular subspaces of $Z$ corresponding to $\mathcal{I}^c$ are $\delta$-close to some singular subspaces perpendicular to $U$ and $V$. Denote them as $\widetilde{U}$ and $\widetilde{V}$ respectively. Then we obtain
    \begin{align*}
	    &\|P_{U_z(:,\mathcal{I}^c)} - P_{U_1}\| = \mathcal{O}(\delta), \qquad \|P_{V_z(:,\mathcal{I}^c)} - P_{V_1}\| = \mathcal{O}(\delta)
	 \end{align*}
	 Let $\widetilde{B}= U_z(:,\mathcal{I}^c)$ and $\widetilde{C} = V_z(:,\mathcal{I}^c)$ and we have the full result. Note that by Lemma \ref{lem: sin_theta}, the constant in the $\mathcal{O}(\cdot)$ notation only depends on the gap between the two groups of singular values, which in our case is determined by the smallest singular value of $X$.
\end{proof}	 
    
\subsection{Proof of Lemma \ref{lem: ness}}
\begin{proof}
Following the proof of Lemma \ref{lem: balls}, we have $U=U_zR+\widetilde{U}_zQ = U_z (Q_1(:,[s]),\, Q_2(:,[r-s]))+\widetilde{U}_z(Q_3(:,[s]), \,Q_4(:,[r-s]))$. Therefore, $R=(Q_1(:,[s]),\,Q_2(:,[r-s]))$, and we have $$R=\begin{pmatrix}I_s+E_1&E_2\\E_3&E_4 \end{pmatrix},$$
with $\|E_1\|^{\frac{1}{2}}$,$\|E_3\|$, $\|E_2\|$, $\|E_4\|\leq\mathcal{O}(\delta)$.
\end{proof}

\subsection{Proof of Lemma \ref{lem: LS}}
\begin{proof}
The proof is a direct consequence of Theorem \ref{thm: LD2converge} and Lemma \ref{lem: P_Tnorm}. Specifically, to make use of Theorem \ref{thm: LD2converge}, it suffices to check Condition (\ref{eq: L}) with  $\omega=\frac{1}{2}$ and Condition (\ref{eq: D}). 

As stated in the beginning of section \ref{sec: DLconverge}, we assume $\{Z_k\}$ is bounded, i.e. $\|Z\|_F\leq C$. In the region $\{Z:\|P_{T_z}(Z-X)\|\geq C_L\}$, we have $\frac{\|P_{T_z}(Z-X)\|_F}{\|Z-X\|_F}\geq \frac{C_L}{C+\|X\|_F}$; on the other hand, in the region $\{Z:\|Z-X\|_F\leq\frac{d_r}{2}\}$, by Lemma \ref{lem: Hoff-W} and Lemma \ref{lem: P_Tnorm}, $\frac{\|P_{T_z}(Z-X)\|_F}{\|Z-X\|_F}\geq \frac{d_r^2}{d_r^2+4\|X\|_F^2}$. One can take $C_1=\min\left\{\frac{C_L}{C+\|X\|_F},\frac{d_r^2}{d_r^2+4\|X\|_F^2}\right\}\geq \Omega(C_L)>0$. In other words, in the region $\{Z:\|P_{T_z}(Z-X)\|\geq C_L\}\cup\{Z:\|Z-X\|_F\leq\frac{d_r}{2}\}$, for $F_1(Z)=\frac{1}{2}\|Z-X\|_F^2$, we have condition (L) holds with $\omega=\frac{1}{2}$ and $C_l=\max\left\{\frac{C+\|X\|_F}{C_L},1+\frac{4\|X\|_F^2}{d_r^2}\right\}$.

For condition (\ref{eq: D}), we consider
\begin{align*}
    f_k-f_{k+1}&=\frac{1}{2}\|Z_k-X\|_F^2-\frac{1}{2}\|Z_{k+1}-X\|_F^2\\
    &=\frac{1}{2}\text{tr}((Z_k+Z_{k+1}-2X)(Z_k-Z_{k+1}))\\
    &=\text{tr}((X-Z_k)(Z_{k+1}-Z_k))-\frac{1}{2}\|Z_{k+1}-Z_k\|_F^2.
\end{align*}
Let $Z_{k+1}=Z_k+\alpha\widetilde{\xi}_k$ and $\xi_n=-P_{T_{Z_k}}(\grad f(Z_k))$, then by first-order retraction property we have $\widetilde{\xi}_k=\xi_k+o(\|\xi_k\|_F)$. So the left- and right-hand side of Condition (\ref{eq: D}) are
\begin{align*}
    \text{LHS} &= f_k-f_{k+1} \\
    &=\text{tr}(-\grad f(Z_k)\alpha\widetilde{\xi}_k)-\frac{\alpha^2}{2}\|\widetilde{\xi}_k\|_F^2\\
    &=\text{tr}(-\grad f(Z_k)(-\alpha P_{T_{Z_k}}(\grad f(Z_k))+o(\alpha\|\xi_k\|_F)))-\frac{\alpha^2}{2}\|\widetilde{\xi}_k\|_F^2\\
    &=\alpha \|\xi_k\|_F^2+o(\alpha\|\xi_k\|_F),\\
    \text{RHS} &= C_d \|P_{T_{Z_k}}(\grad f(Z_k))\|_F\|Z_{k+1}-Z_k\|_F  \\
    &=C_d \|\xi_k\|_F\|\alpha\widetilde{\xi}_k\|_F \\ &=C_d\alpha\|\xi_k\|_F^2+o(\alpha\|\xi_k\|_F^2).
\end{align*}
By choosing a proper $C_d>0$ and a small enough step size $\alpha$, one can get $$ f_k-f_{k+1}\geq  C_d \|P_{T_{z_k}}(\grad f(Z_k))\|_F \| X_{k+1}-X_k\|_F,$$ which is Condition (\ref{eq: D}). The results now follow from Theorem \ref{thm: LD2converge}.
\end{proof}

\subsection{Proof of Lemma \ref{lem: uniformST}}
\begin{proof}
Since the marginal distribution of W(:,i) (i=1,2,...,r) is the  uniform distribution on $\mathcal{S}^{n-1}$ and the distribution of $W$ is right-rotational invariant, we assume unitary $U\in\mathbb{F}^{n\times n}$ such that $Uu_0=e_1$ and $UW=\widetilde{W}$ and use the distribution of $\widetilde{W}$ to replace that of $W$. Now consider the marginal distribution of $W_i$, and write $W_i=\frac{g}{\|g\|}$ where $g$ is drawn from $\mathcal{N}(0,I_n)$. Then, by the Bernstein-type inequality, we have the following estimation:
\begin{align*}
    \text{Prob}\left(\left|\|g\|^2-\mathbb{E}(\|g\|^2)\right|>t\mathbb{E}(\|g\|^2)\right)\leq 2\exp\left\{-c\min\left\{\frac{t^2\mathbb{E}(\|g\|^2)^2}{K^2n},\frac{t\mathbb{E}(\|g\|^2)}{K}\right\}\right\}.
\end{align*}
This implies $\text{Prob}\left(\|g\|^2\in(\frac{1}{2}n,\frac{3}{2}n)\right)\geq 1-e^{-\Omega(n)}$. On the other hand, we have
\begin{align*}
    \text{Prob}\left(|g(1)|^2-1>t\right)&=\text{Prob}(|g(1)|>\sqrt{1+t})\\
    &=2\int_{\sqrt{1+t}}^{\infty}\frac{1}{\sqrt{2\pi}}e^{-\frac{x^2}{2}}dx=:\sqrt{\frac{2}{\pi}}A, \\
    \text{where}\qquad A^2&=\int_{\sqrt{1+t}}^{\infty}\int_{\sqrt{1+t}}^{\infty}e^{-\frac{x^2+y^2}{2}}dxdy=\frac{\pi}{2} e^{-\frac{1+t}{2}}.
\end{align*}
Taking $t=-1+c_1\frac{1}{\log n}$, we have
$$\text{Prob}\left(|g(1)|^2>c_1\frac{1}{\log n}\right)=e^{-{c_1}/({4\log n})}\geq 1-\mathcal{O}(\frac{1}{\log n}).$$
Therefore, we have $|u_0^{\dag}W_i|^2\gtrsim \frac{1}{n\log n}$ with the fail probability controlled by $e^{-\Omega(n)}+\mathcal{O}(\frac{1}{\log n})$. By taking $t=c_1\log n -1$, with $c_1=\Omega(1)$, we have:
\begin{align*}
    \text{Prob}(|g(1)|^2\geq c_1\log n )\leq e^{-c_1\log(n) / 4}=\frac{1}{n^{{c_1}/{4}}}.
  \end{align*}
Therefore, we have $|u_0^{\dag}W_i|^2\lesssim\frac{\log n}{n}$ holds, with the fail probability controlled by $e^{-\Omega(n)}+\frac{1}{\text{poly}(n)}$. Taking  $t=-1+{c_2}/{n^p}$, we have 
\begin{align*}
    \frac{|g(1)|^2}{\|g\|^2}\gtrsim
    \frac{1}{n^{p+1}}
\end{align*}
holds with fail probability controlled by $e^{-\Omega(n)}+\mathcal{O}(\frac{1}{n^p})$.
For the complex case, the proof is very similar and we omit the details here.
\end{proof}

\subsection{Proof of Lemma \ref{lem: Rdynamics}}
\begin{proof}
\begin{enumerate}[label=\arabic*)]
    \item Denote $Z=U_zS_zU_z^{\dag}$ as an alternative decomposition of $Z$ with $U_z^{\dag}U_z=I_r$ and $S_z\in\R^{r\times r}$. Assume $P\in O(r)$ such that: $V_z=U_zP$ and $P^{\dag}S_zP=\Sigma_z$. Denote $U=U_zR_*+\widetilde{U}_zQ_*=V_zR+\widetilde{V}_zQ$. By applying the dynamical low-rank approximation in Lemma \ref{lem: dylowrank}, we have:
\begin{align*}
    &\frac{d}{dt}U_z=\widetilde{U}_zQ_*DR_*^{\dag}S_z^{-1}\\
    &\frac{d}{dt}S_z=R_*DR_*^{\dag}-S_z.
\end{align*}
Therefore, we obtain
\begin{align*}
    \frac{d}{dt}R_*&=\frac{d}{dt}(U_z^{\dag}U)=S_z^{-1}R_*DQ_*^{\dag}Q_*.
\end{align*}
Using $R=V_z^{\dag}U=P^{\dag}U_z^{\dag}U=P^{\dag}R_*$, we have

\begin{equation}\label{eq: Rdy}
\begin{split}
    \frac{d}{dt}R&=\frac{d}{dt}\left(P^{\dag}R_*\right)\\
    &=\frac{d}{dt}(P)^{\dag} R_*+P^{\dag}\frac{d}{dt}(R_*)\\
    &=\frac{d}{dt}(P)^{\dag}PR+P^{\dag}S_z^{-1}PRDQ_*^{\dag}Q_*\\
    &=\left(\frac{d}{dt}(P)^{\dag}P\right)R+\Sigma_z^{-1}RDQ^{\dag}Q.
\end{split}
\end{equation}
The last equation follows from $P^{\dag}S_z^{-1}P=\Sigma_z^{-1}$ and $Q^{\dag}Q=I_r-R^{\dag}R=I_r-R^{\dag}P^{\dag}PR=Q_*^{\dag}Q_*$. Then, we have
\begin{equation}
\begin{split}
    \frac{d}{dt}(RDR^{\dag})&=\Sigma_z^{-1}RDQ^{\dag}QDR^{\dag}+RDQ^{\dag}QDR^{\dag}\Sigma_z^{-1}\\
    &+\left(\frac{d}{dt}(P)^{\dag}P\right)RDR^{\dag}+RDR^{\dag}\left(P^{\dag}\frac{d}{dt}(P)\right). 
    \end{split}
\end{equation}

Due to the fact that $P^{\dag}P=I_r$, we have $\frac{d}{dt}(P)^{\dag}P+P^{\dag}\frac{d}{dt}(P)=0$. Denote $M=P^{\dag}\frac{d}{dt}(P)$, then $M$ is an antisymmetric matrix such that $M+M^{\dag}=0$. 
Therefore, we have
\begin{align*}
    &\frac{d}{dt}\left(RDR^{\dag}\right)=M^{\dag}RDR^{\dag}+RDR^{\dag}M+\Sigma_z^{-1}RDQ^{\dag}QDR^{\dag}+RDQ^{\dag}QDR^{\dag}\Sigma_z^{-1}.
 \end{align*}

On the other hand, we have
\begin{align*}
    \frac{d}{dt}(\Sigma_z)&=\frac{d}{dt}(P^{\dag}S_zP)\\
    &=\frac{d}{dt}(P)^{\dag}S_zP+P^{\dag}\frac{d}{dt}(S_z)P+P^{\dag}S_z\frac{d}{dt}(P)\\
    &=P^{\dag}\left(R_*DR_*^{\dag}-S_z\right)P+\left(\frac{d}{dt}(P)^{\dag}P\right)\Sigma_z+\Sigma_z\left(P^{\dag}\frac{d}{dt}(P)\right)\\
    &=RDR^{\dag}-\Sigma_z+\left(\frac{d}{dt}(P_z)^{\dag}P_z\right)\Sigma_z+\Sigma_z\left(P_z^{\dag}\frac{d}{dt}(P_z)\right).
\end{align*}
Thus, we arrive at
\begin{equation}
\begin{split}
    \frac{d}{dt}(\Sigma_z(s,s))&=(RDR^{\dag})(s,s)-\Sigma_z(s,s).
    \end{split}
\end{equation}
Since $\frac{d}{dt}(\text{offdiag}(\Sigma_z))=0$, we have 
\begin{align}
    \label{eq: M}
    \text{offdiag}(RDR^{\dag})=-M^{\dag}\Sigma_z-\Sigma_zM=M\Sigma_z+\Sigma_zM^{\dag}.\footnote{Here, $\text{offdiag}(A):=A-diag(A)$ which keeps the off-diagonal entries of $A$.} 
\end{align}

By substitution, we get
\begin{align}
    \frac{d}{dt}(RDR^{\dag})(j,j)&=e_j^{\dag}(2M\Sigma_zM-\Sigma_zMM-MM\Sigma_z)e_j+\frac{2}{\sigma_j(Z)}\|QDR^{\dag}e_j\|_2^2\nonumber\\
    &=2\sum_{k=1}^r(\sigma_j(Z)-\sigma_k(Z))M(k,j)^2+\frac{2}{\sigma_j(Z)}\|QDR^{\dag}e_j\|_2^2 \\ 
    &=2\sum_{k\neq j}\frac{1}{\sigma_j-\sigma_k}(RDR^{\dag})(k,j)^2+\frac{2}{\sigma_j(Z)}\|QDR^{\dag}e_j\|_2^2.\nonumber
\end{align}
Note that
\begin{align*}
    \sum_{j=1}^r\sum_{k=1}^r (\sigma_j(Z)-\sigma_k(Z))M(k,j)^2&=-\sum_{k=1}^r\sum_{j=1}^r(\sigma_k(Z)-\sigma_j(Z))M(k,j)^2\\
    &=-\sum_{k=1}^r\sum_{j=1}^r(\sigma_j(Z)-\sigma_k(Z))M(k,j)^2.
\end{align*}
Thus, we obtain
\begin{align}\label{eq: RDR} 
    \frac{d}{dt}(\sum_{j=1}^r \|D^{\frac{1}{2}}R^{\dag}e_j\|^2)&=\sum_{j=1}^r \frac{2}{\sigma_j(Z)}\|QDR^{\dag}e_j\|^2.
\end{align}

\item Denote $R=T_2\Sigma_{RR}T_1^{\dag}$. Simple calculation gives $R^{\dag}R=T_1\Sigma_{RR}T_1^{\dag}$ and $Q^{\dag}Q=T_1\Sigma_{QQ}T_1^{\dag}$ with $\Sigma_{RR}+\Sigma_{QQ}=I_r$. We have
\begin{align*}
    \sum_{j=1}^r(RDR^{\dag})(j,j)&=\text{trace}(RDR^{\dag})=\text{trace}(DR^{\dag}R)\\
    &=\text{trace}(T_1^{\dag}DT_1\Sigma_{RR})\\
    &=\sum_{j=1}^r t_jr_j.
\end{align*}
Here, for each $j\in[r]$, $t_j=\|D^{\frac{1}{2}}T_1e_j\|^2=\Omega(1)$ and $\Sigma_{RR}(j,j)=r_j$. Now, for a bounded sequence $\{Z_t\}$ we have
\begin{align*}
    \sum_{j=1}^r\frac{2}{\sigma_j}\|QDR^{\dag}e_j\|^2&\gtrsim \text{trace}(RDQ^{\dag}QDR^{\dag})\\
    &=\text{trace}(T_1^{\dag}DT_1\Sigma_{QQ}T_1^{\dag}DT_1\Sigma_{RR}).
\end{align*}
Denote $N=T_1^{\dag}DT_1$. We obtain the following estimate:
\begin{align*}
    \sum_{j=1}^r\frac{2}{\sigma_j}\|QDR^{\dag}e_j\|^2
    & \gtrsim \sum_{j=1}^r r_j(1-r_j)N(j,j)^2\\
    &\gtrsim\sum_{j=1}^r r_j(1-r_j).\end{align*}
This gives
\begin{align*}
    \frac{d}{dt}(\sum_{j=1}^r r_j)\gtrsim \sum_{j=1}^r r_j(1-r_j).
\end{align*}
Further, if $\sigma_j=\Omega(1)$, for all $j\in[r]$, by equation (\ref{eq: RDR}), we have
\begin{align*}
    \frac{d}{dt}(\sum_{j=1}^r \|D^{\frac{1}{2}}R^{\dag}e_j\|^2)&= \Omega\left(\sum_{j=1}^r \|QDR^{\dag}e_j\|^2\right)\\
    &=\Omega\left(\text{trace}\left(T_1^{\dag}DT_1\Sigma_{QQ}T_1^{\dag}DT_1\Sigma_{RR}\right)\right)\\
    &=\Omega\left(\sum_{j=1}^r r_j(1-r_j)\right).
\end{align*}

Thus we have $\frac{d}{dt}\sum_{j=1}^r r_j =\Omega\left(\sum_{j=1}^r (1-r_j)r_j\right)$.
\end{enumerate}
\end{proof}

\subsection{Proof of Lemma \ref{lem: P_Tnorm}}
\begin{proof}
For simplicity, the following proof is based on the symmetric case where $Z = U_z \Sigma U_z^\dag$ and $X = U D U^\dag$. 
\begin{enumerate}[label=\arabic*)]
    \item
    Let $\widetilde{U}_z \in \F^{n\times (n-r)}$ be the orthogonal complement of $U_z$, and 
    \begin{align*}
        X  = (U_z, \widetilde{U}_z)
        \widetilde{X} (
            U_z ,
            \widetilde{U}_z
        )^\dag
        =  (U_z, \widetilde{U}_z)
        \begin{pmatrix}
            X_{11} & X_{12} \\
            X_{21} & X_{22}
        \end{pmatrix} 
        \begin{pmatrix}
            U_z^\dag \\
            \widetilde{U}_z^\dag
        \end{pmatrix}.
    \end{align*}
    
Then, we get 
    \begin{align*}
		\|P_{T_z}(Z-X)\|_F &= \left\|
		\begin{pmatrix}
			X_{11}-\Sigma_z & X_{12} \\
			X_{21} & 0
		\end{pmatrix}\right\|_F = \|X_{11}-\Sigma_z\|_F^2 + \|X_{12}\|_F^2 + \|X_{21}\|_F^2, \\
		\|Z-X\|_F^2 &= \|P_{T_z}(Z-X)\|_F^2 + \|X_{22}\|_F^2.
	\end{align*}
We also assume that $U = U_z R + \widetilde{U}_z S$, where $R \in\F^{r\times r}$ and $S \in \F^{(n-r)\times r}$. Then, we obtain
\begin{align*}
    X_{11} = R D R^\dag, \quad X_{12} = R D S^\dag, \quad X_{21} = S D R^\dag, \quad X_{22} = S D S^\dag.
\end{align*}
The goal is to find lower and upper bounds for
\begin{align*}
    s = \frac{\|P_{T_z}(Z-X)\|_F^2}{\|Z-X\|_F^2}.
\end{align*}
It is obvious that $s\le 1$. 
To identify the lower bound we consider 
\begin{align*}
    \phi = \frac{\|Z-X\|_F^2}{\|X_{22}\|_F^2}, \quad s = \frac{\phi-1}{\phi}, \quad \phi\ge 1.
\end{align*}
For fixed $\Sigma$ and $D$, $\phi$ can be seen as a function of $R$. We express $\phi(R)$ in terms of $\phi_1$ and $\phi_2$ defined below:
\begin{align*}
    \phi(R) = \frac{\phi_1(R)}{\phi_2(R)}, &\quad R \in \mathcal{D}\subset \F^{r\times r}, \, \mathcal{D} = \{R : 0\preccurlyeq R^{\dag}R\preccurlyeq I_r\}, \\
     \text{where }\phi_1(R) &= \|X\|_F^2 + \|\Sigma\|_F^2 - 2\langle\Sigma, X_{11} \rangle \\
    & = \|D\|_F^2 + \|\Sigma\|_F^2 - 2\text{tr}(\Sigma RD R^{\dag}), \\
    \phi_2(R) &= \text{tr}\left((S DS^{\dag})^2 \right)= \text{tr}\left((DS^{\dag}S)^2 \right)\\
    &= \text{tr}\left((D(I-R^{\dag}R))^2\right).
\end{align*}
To minimize $t$ for given $D$ and $\Sigma$, the first-order condition is obtained by taking derivative of $t$ over $R$:
\begin{align*}
    \frac{\partial t}{\partial R} &= \frac{1}{\phi_2^2}(\frac{\partial \phi_1}{\partial R}\phi_2 - \frac{\partial \phi_2}{\partial R}\phi_1) \\
    &= \frac{1}{\phi_2^2}(-4\phi_2\Sigma RD + 4\phi_1RD(I-R^{\dag}R)D) \\ 
    &= \frac{4}{\phi_2^2}(-\phi_2\Sigma RD + \phi_1RDS^{\dag}S D).
\end{align*}
Imposing $\partial t/\partial R = 0$ gives
\begin{align}
\label{eq: first-order}
    \Sigma RD = \phi\cdot RDS^{\dag}S D.
\end{align}
We now claim that this first-order condition \emph{cannot} be satisfied in the interior of the domain $\mathcal{D}$. To see this, note that the above equation (\ref{eq: first-order}) gives
\begin{align*}
    \Sigma RD R^{\dag} = \phi\cdot RDS^{\dag}S D R^{\dag},\quad \text{ i.e. }\quad \langle \Sigma, X_{11}\rangle = \phi\|X_{12}\|_F^2 = \phi\|X_{21}\|_F^2.
\end{align*}
Thus, we have
\begin{align*}
    \phi &= \frac{\|Z-X\|_F^2}{\|X_{22}\|_F^2} = \frac{\|\Sigma\|_F^2+\|X_{11}\|_F^2 - 2\langle \Sigma, X_{11}\rangle + \|X_{12}\|_F^2 + \|X_{21}\|_F^2 +\|X_{22}\|_F^2}{\|X_{22}\|_F^2} \\
    &= \frac{\|\Sigma\|_F^2 +\|X_{11}\|_F^2 - (2-2/\phi)\langle \Sigma, X_{11}\rangle}{\|X_{22}\|_F^2}+1 \\
    &= \frac{\|X_{11}-(1-1/\phi)\Sigma\|_F^2 }{\|X_{22}\|_F^2} + (\frac{2}{\phi}-\frac{1}{\phi^2})\frac{\|\Sigma\|_F^2}{\|X_{22}\|_F^2} +1 \\
    &\ge (\frac{2}{\phi}-\frac{1}{\phi^2})\frac{\|\Sigma\|_F^2}{\|X_{22}\|_F^2} +1.
\end{align*}
Note that when $R \in int(\mathcal{D})$, $R$ is full-rank. Since $D, \Sigma$ are also full-rank, (\ref{eq: first-order}) gives $\phi^2 = \|\Sigma\|_F^2/\|X_{22}\|_F^2$ by some simple matrix manipulation: 
    \begin{align*}
        & \Sigma RD = \phi\cdot RDS^{\dag}S D \\
        \Rightarrow \quad & R^{-1}\Sigma R = \phi\cdot DS^{\dag}S \\
        \Rightarrow \quad & \text{tr}\left((R^{-1}\Sigma R)^2\right) = \phi^2 \text{tr} \left((DS^{\dag}S)^2\right)\\
        \Rightarrow \quad & \text{tr}\left(\Sigma^2\right) = \phi^2 \text{tr} \left((S DS^{\dag})^2\right) \\
        \Rightarrow \quad & \|\Sigma\|_F^2 = \phi^2\|X_{22}\|_F^2.
    \end{align*}
Hence, we obtain
\begin{align*}
    \phi \ge  (\frac{2}{\phi}-\frac{1}{\phi^2})\phi^2 +1  = 2\phi-1+1 = 2\phi,
\end{align*}
which contradicts the fact that $\phi\ge 1$.

Therefore, either the extreme values of $t$ are only achieved on $\partial \mathcal{D}$, or the full-rankness of $D$ is violated. In either case, the RHS of (\ref{eq: first-order}) is rank-deficient. So the LHS of (\ref{eq: first-order}) (and thus $R$ and $X_{11}$) is also rank-deficient. Therefore, we conclude that 
\begin{align*}
    \phi-1 \ge \frac{\|X_{11}-\Sigma\|_F^2}{\|X_{22}\|_F^2} \ge \frac{\|X_{11}-\Sigma\|_F^2}{\|X\|_F^2} \ge \frac{\sigma_r^2}{\|X\|_F^2},
    \end{align*}
where $\sigma_r$ is the smallest nonzero eigenvalue of $Z$. Finally, we obtain
\begin{align*}
    s \ge \frac{\sigma_r^2}{\sigma_r^2 + \|X\|_F^2}.
\end{align*}
Now we have $\|P_{T_z}(Z-X)\|_F^2 \ge s \|Z-X\|_F^2$ for $s>0$ as long as $Z$ is of rank $r$ and the smallest nonzero singular value of $Z$ is bounded away from 0.
\item By Lemma \ref{lem: Hoff-W}, when $\|Z-X\|_F<\frac{d_r}{2}$, we have $\sigma_r>\frac{d_r}{2}$. Further, by (1) we have $\frac{\|P_{T_z}(Z-X\|_F}{\|Z-X\|_F}\gtrsim d_r$. By Lemma \ref{lem: LS}, we have the locally linear convergence towards $X$ with $c=-\log\left(1-\Omega(\alpha d_r^2)\right)$. To prove the result for single-step convergence, we consider
\begin{align*}
    f_{k+1}&\leq f_k-C_2\|P_{T_z}(Z_k-X)\|_F\|Z_{k+1}-Z_k\|_F\\
    &\leq f_k-C_2(\alpha+o(\alpha))\cdot \|P_{T_z}(Z_k-X)\|_F^2\\
    &\leq (1-\Omega(\alpha)C_1^2C_2)f_k.
\end{align*}
Here, $f_k:=\frac{1}{2}\|Z_k-X\|_F^2$. The first inequality is from condition (D). The second inequality is by the first-order retraction, and the third inequality is from condition (L). Therefore, we have
\begin{align*}
    \|Z_{k+1}-X\|_F\leq \sqrt{1-\Omega(\alpha)C_1^2C_2}  \|Z_k-X\|_F.
\end{align*}
That is, $\|Z_{k+1}-X\|_F\leq e^{-c}\|Z_k-X\|_F$, with $c=-\log(1- \Omega\left(\alpha d_r^2)\right)$.
\end{enumerate}
\end{proof}

\subsection{Proof of Theorem \ref{thm: Converge12}}
\begin{proof} 
The proof of Theorem \ref{thm: Converge12} can be reduced to deducing Conditions (\ref{eq: L}) and (\ref{eq: D}) from Conditions (1) and (2) in the assumptions. Note that in the setting of  $T:\mathcal{M}\rightarrow \R^m$, $T(Z)=\frac{1}{\sqrt{m}}\begin{pmatrix}\langle A_1,Z\rangle\\ \langle A_2,Z\rangle\\...\\\langle A_m,Z\rangle\end{pmatrix} $, $f(Z)=\frac{1}{2}\|T(Z)-T(X)\|_2^2$, Conditions (\ref{eq: L}) and (\ref{eq: D}) can be formulated as follows.
\begin{enumerate}[label=\arabic*)]
    \item Condition (\ref{eq: L}): {\L}ojasiewicz gradient inequality
    $$
        \left(\frac{1}{2m}\sum_j\langle A_j,Z-X\rangle ^2\right)^{1-\omega}\leq C_l\|P_{T_z}( \frac{1}{m}\sum_j\langle A_j,Z-X\rangle A_j)\|_F
    $$
    holds with $\omega=\frac{1}{2}$.
    \item Condition (\ref{eq: D}): 
    $$
        f_k-f_{k+1}\geq C_d \|P_{T_{Z_k}}(\frac{1}{m}\sum_j\langle A_j,Z_k-X\rangle A_j )  \|_F\|Z_{k+1}-Z_k\|_F.
    $$
\end{enumerate}

The proof now goes as follows.
\begin{enumerate}[label=\arabic*)]
    \item To prove condition (\ref{eq: L}): 
    By (1) and (2) we have
    \begin{align*}
        \|P_{T_z}(\grad f(Z))\|_F&\geq \frac{1}{C_3}\|Z-X\|_F\\
        &\geq \frac{1}{C_2C_3}\|T(Z)-T(X)\|\\
        &=\frac{\sqrt{2}}{C_2C_3}\left|f(Z)-f(X)\right|^{\frac{1}{2}},
    \end{align*}
    which implies that (\ref{eq: L}) holds with $\omega=\frac{1}{2}$ and $C_l=\frac{\sqrt{2}C_2C_3}{2}>0$ is an abosolute constant.
    \item To prove condition (\ref{eq: D}): we first consider
    \begin{align*}
        f_k-f_{k+1}&=\frac{1}{2}\|T(Z_k)-T(X)\|_2^2-\frac{1}{2}\|T(Z_{k+1})-T(X)\|_2^2\\
        &=\frac{1}{2}\langle T(Z_k+Z_{k+1}-2X), T(Z_k-Z_{k+1}) \rangle\\
        &=\langle T^*T(X-Z_k), Z_{k+1}-Z_k\rangle-\frac{1}{2}\|T(Z_{k+1}-Z_k)\|_F^2.
    \end{align*}
    Assume that $Z_{k+1}=Z_k+\alpha\widetilde{\xi}_k$, and $-\alpha P_{T_{Z_k}}(\grad f(Z_k))=\alpha\xi_k$. By first-order retraction property and condition (1), we get
    \begin{align*}
        f_k-f_{k+1}&\geq \langle -\grad f(Z_k),\alpha\widetilde{\xi}_k\rangle-\frac{C_2^2\alpha^2}{2}\|\widetilde{\xi}_k\|^2\\
        &=\langle -\grad f(Z_k),\alpha \xi_k\rangle+o(\alpha\|\xi_k\|_F^2)\\
        &=\langle -\grad f(Z_k),-P_{T_{Z_k}}(\grad f(Z_k))\rangle+o(\alpha\|\xi_k\|_F^2)\\
        &=\alpha\|P_{T_{Z_k}}(\grad f(Z_k))\|_F^2+o(\alpha\|\xi_k\|_F^2).
    \end{align*}
    On the other hand, we also obtain
    \begin{align*}
        C_d \|P_{T_{Z_k}}(\grad f(Z_k))\|_F\|Z_{k+1}-Z_k\|_F
        &=C_d\|P_{T_{Z_k}}(\grad f(Z_k))\|_F \|\alpha\widetilde{\xi}_k\|_F\\
        &=C_d\alpha\|P_{T_{Z_k}}(\grad f(Z_k))\|_F^2+o(\alpha\|\xi_k\|_F^2).
    \end{align*}
    By choosing $C_d>0$ small enough, we have
    $$
        f_k-f_{k+1}\geq C_d \|P_{T_{Z_k}}(\frac{1}{m}\sum_j\langle A_j,Z_k-X\rangle A_j )  \|_F\|Z_{k+1}-Z_k\|_F,
    $$
    i.e. Condition (\ref{eq: D}) holds.
\end{enumerate}
From Theorem \ref{thm: LD2converge}, we conclude that projected gradient descent for the least squares loss function $f(Z)=\frac{1}{2}\|T(Z)-T(X)\|_2^2$ converges to its global minimum linearly.
\end{proof}

\section{Auxiliary Lemmas}\label{sec: auxLM}
\begin{lemma}[The $\text{sin}(\theta)$ Theorem,  \cite{davis1970rotation}]
\label{lem: sin_theta}
    Let $A$ be a Hermitian operator. Assume that
    \begin{align*}
        A = 
        \begin{pmatrix}
            E_0 & E_1
        \end{pmatrix}
        \begin{pmatrix}
            A_0 & 0 \\
            0 & A_1
        \end{pmatrix}
        \begin{pmatrix}
            E_0^H \\
            E_1^H
        \end{pmatrix}
    \end{align*}
    is an invariant subspace decomposition (i.e. generalized eigenvalue decomposition) of $A$. Let
    \begin{align*}
        B = A + \Delta, \quad B = 
        \begin{pmatrix}
            F_0 & F_1
        \end{pmatrix}
        \begin{pmatrix}
            B_0 & 0 \\
            0 & B_1
        \end{pmatrix}
        \begin{pmatrix}
            F_0^H \\
            F_1^H
        \end{pmatrix}.
    \end{align*}
    Let $\Theta_0$ be the angle matrix between subspaces $E_0$ and $F_0$. Define the residual as 
    \begin{align*}
        R = BE_0 - E_0A_0.
    \end{align*}
    If there is an interval $[\beta, \alpha]$ and $\delta >0$, such that the spectrum of $A_0$ lies entirely in $[\beta, \alpha]$, while that of $B_1$ lies entirely in $(-\infty, \beta-\delta] \cup [\alpha+\delta, +\infty)$, then for every unitary-invariant norm, we have
    \begin{align*}
        \delta \|\text{sin}\Theta_0\| \le \|R\|.
    \end{align*}
    In particular, this holds true for the matrix 2-norm and the Frobenius norm. 
\end{lemma}

\begin{lemma}[Hoffman-Wielandt Theorem]
\label{lem: Hoff-W}
\leavevmode
\begin{enumerate}[label=\arabic*)] 
    \item Assume $Z,Z'\in\R^{n\times n}$ or $\C^{n\times n}$ are normal matrices, and their corresponding ordered spectra are $\{\lambda_j\}$ and $\{\tilde{\lambda_j}\}$. Then, we have
    \begin{align*}
        \sqrt{\sum_{j=1}^n|\tilde{\lambda_j}-\lambda_j|^2}\leq \|Z'-Z\|_F.
    \end{align*}
    \item Assume $Z$, $Z'\in \R^{m\times n}$ or $\C^{m\times n}$, and denote their singular values in descending order as $\{\sigma_j\}$ and $\{\tilde{\sigma_j}\}$. Then, we have
        \begin{align*}
        \sqrt{\sum_{j=1}^n|\tilde{\sigma_j}-\sigma_j|^2}\leq \|Z'-Z\|_F.
    \end{align*}
\end{enumerate}
\end{lemma}

\begin{lemma}[Dynmical low-rank approximation, \cite{DyLowRank}]
\label{lem: dylowrank}
Assume $Z=USV^{\dag}$ with $M:=P_{T_{Z}}(\frac{d}{dt}g(Z))=\grad_{\M_r}g(Z)$, then gradient flow satisfies:
\begin{align*}
    &\frac{d}{dt}S=U^{\dag}MV,\\
    &\frac{d}{dt}U=P_U^{\perp}MVS^{-1},\\
    &\frac{d}{dt}V=P_V^{\perp}M^{\dag}US^{-\top}.
\end{align*}
Here, $P_U^{\perp}=I-UU^{\dag}$ and $P_V^{\perp}=I-VV^{\dag}$.
\end{lemma}
\end{document}